\documentclass[10pt,twocolumn,letterpaper]{article}
\usepackage[pagenumbers]{cvpr}

\usepackage{amsmath,amsfonts,bm}

\def\vc{{\bm{c}}}

\def\vx{{\bm{x}}}
\def\vy{{\bm{y}}}

\def\gX{{\mathcal{X}}}

\definecolor{cvprblue}{rgb}{0.21,0.49,0.74}
\usepackage[pagebackref,breaklinks,colorlinks,allcolors=cvprblue]{hyperref}
\def\paperID{18403} 
\def\confName{CVPR}
\def\confYear{2025}

\usepackage{comment}
\usepackage{enumitem}
\usepackage{graphicx}
\usepackage{comment}
\usepackage{booktabs}
\usepackage{multirow}
\usepackage{rotating}
\usepackage{pifont}
\usepackage{threeparttable}
\usepackage{makecell}
\usepackage{xcolor}
\usepackage{xurl}
\usepackage{array, booktabs}
\usepackage{colortbl}
\usepackage{amssymb}
\usepackage{stfloats}
\usepackage{float}
\usepackage{stackengine}
\usepackage{colortbl}
\robustify{\bfseries}
\makeatletter
\robustify\@latex@warning@no@line
\makeatother
\usepackage{authblk}  

\makeatletter
\renewcommand\AB@affilsepx{~~ \protect\Affilfont}
\makeatother

\definecolor{deepred}{RGB}{139,0,0} 
\definecolor{lightgray}{rgb}{0.9,0.9,0.9}

\newcommand{\ours}{Align-Anything}

\title{Align Anything: Training All-Modality Models to Follow Instructions with Language Feedback}

\author[1*]{Jiaming~Ji}
\author[1*]{Jiayi~Zhou}
\author[1*]{Hantao~Lou}
\author[1*]{Boyuan~Chen}
\author[1*]{Donghai~Hong}
\author[1]{Xuyao~Wang}
\author[1]{Wenqi~Chen}
\author[1]{Kaile~Wang}
\author[1]{Rui~Pan}
\author[1]{Jiahao~Li}
\author[1]{Mohan~Wang}
\author[1,2]{Josef~Dai}
\author[1]{Tianyi~Qiu}
\author[1]{Hua~Xu}
\author[3]{Dong~Li}
\author[4]{Weipeng~Chen}
\author[5]{Jun~Song}
\author[5]{Bo~Zheng}
\author[1$\dagger$]{Yaodong~Yang}
\affil[1]{Institute for AI, Peking University} 
\affil[2]{Beijing Academy of Artificial Intelligence (BAAI)\newline}
\affil[3]{Huawei Noah's Ark LAB}
\affil[4]{Baichuan Inc.}
\affil[5]{Taobao \& Tmall Group of Alibaba}

\begin{document}

\twocolumn[{%
\renewcommand\twocolumn[1][]{#1}%
\maketitle

\begin{center}
    \centering
    \captionsetup{type=figure}
    \includegraphics[width=\textwidth]{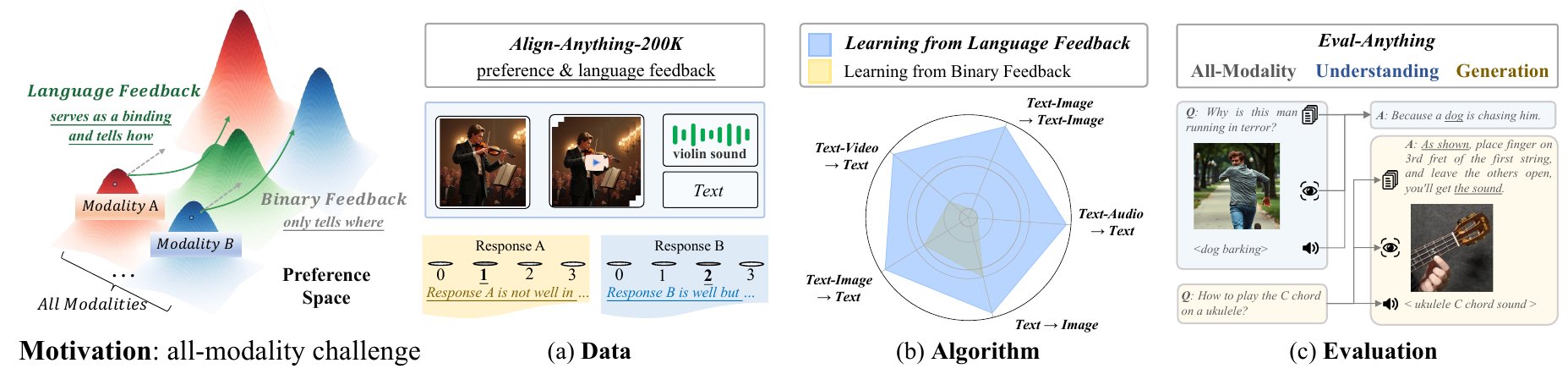}
    \vspace{-2.0em}
    \caption{
    Motivated by the challenge of achieving all-modality human preference alignment, particularly the limitations of binary preferences in accurately reflecting human preferences, we introduce the \textit{align-anything}:  \textbf{Data:} \textit{align-anything-200k}, which covers text, image, audio, video modalities, and 8+ specific subtasks, annotated with preference and language feedback; \textbf{Algorithm:} improving all-modality alignment by learning from language feedback; \textbf{Evaluation:} encompassing all-modality understanding and generation.}
    \label{fig:figure1}
\end{center}%
}]

\begin{abstract}
\renewcommand{\thefootnote}{*}\footnotemark
\footnotetext{Core authors, $^\dag$Corresponding author. This is a collaborative project and if you have any questions, please feel free to reach out via email at \textit{{\{jiamg.ji, gaiejj\}@stu.pku.edu.cn}} or \textit{yaodong.yang@pku.edu.cn.}}
Reinforcement learning from human feedback (RLHF) has proven effective in enhancing the instruction-following capabilities of large language models; however, it remains underexplored in the cross-modality domain. As the number of modalities increases, aligning all-modality models with human intentions -- such as instruction following -- becomes a pressing challenge. In this work, we make the first attempt to fine-tune all-modality models (i.e. input and output with any modality, also named any-to-any models) using human preference data across all modalities (including text, image, audio, and video), ensuring its behavior aligns with human intentions. This endeavor presents several challenges. First, there is no large-scale all-modality human preference data in existing open-source resources, as most datasets are limited to specific modalities, predominantly text and image. Secondly, the effectiveness of binary preferences in RLHF for post-training alignment in complex all-modality scenarios remains an unexplored area. Finally, there is a lack of a systematic framework to evaluate the capabilities of all-modality models, particularly regarding modality selection and synergy. To address these challenges, we propose the align-anything framework, which includes meticulously annotated 200k all-modality human preference data. Then, we introduce an alignment method that learns from unified language feedback, effectively capturing complex modality-specific human preferences and enhancing the model's instruction-following capabilities. Furthermore, to assess performance improvements in all-modality models after post-training alignment, we construct a challenging all-modality capability evaluation framework -- eval-anything. All data, models, and code framework have been open-sourced for the community. For more details, please refer to \url{https://github.com/PKU-Alignment/align-anything}.
\end{abstract}

\section{Introduction}
Our world is inherently multimodal \citep{turk2014multimodal, zhang2023llama, liu2024improved}. Humans perceive the world through various sensory organs, acquiring information in multiple modalities such as text, images, audio, video, and others. 
These different forms of information often complement and interact with each other.
Each sensory channel has unique advantages in conveying specific concepts and enhancing our understanding of the world.
With the success of large language models (LLMs) \citep{achiam2023gpt, anthropic2024claude3, zhao2023survey}, researchers aim to extend these models to handle multiple modalities, enabling them to perceive and generate any modality \citep{alayrac2022flamingo, liu2024visual, team2024chameleon,yu2024rlhf}.
This would allow the models to respond using the most appropriate modality, achieving truly human-like AI \citep{wu2024nextgpt, liu2024llavanext, fang2024llama, yang2024cogvideox, chen2024sharegpt4video, wu2023human,kirstain2023pick,gong2023listen, wang2024vidprom}.

Consequently, the research community has actively developed foundational models capable of handling arbitrary modalities \citep{wu2024nextgpt, li2024uni}.
Based on LLMs, people use individual image/text encoders or domain-specific decoders for input or output processing \citep{zhang2023video, liu2024visual, zhu2024minigpt, wu2024nextgpt}, leveraging the MoE architecture \citep{lin2024moe} and diffusion techniques \citep{ho2020denoising}.
In this line of work, each modality is encoded by a single encoder, with non-text modality information being mapped to the text space via projection layers. 
Additionally, Chameleon \citep{team2024chameleon} has experimented with encoding images during the pre-training phase using fully token-based representations to handle both image and text modalities.
However, handling more modalities remains a significant challenge.

\begin{figure*}[ht]
    \centering
    \includegraphics[width=\textwidth]{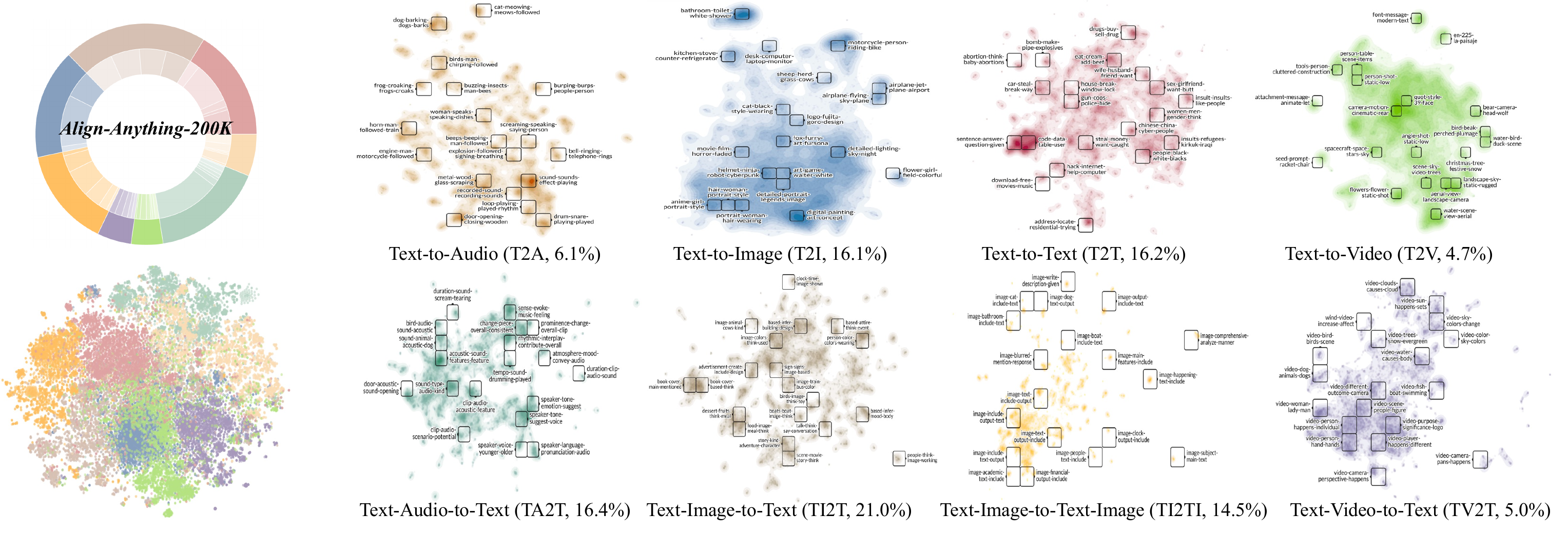}
    \vspace{-2.0em}
    \caption{\textbf{Composition and distribution of \textit{align-anything-200k.}} Our dataset comprises 8 subtasks across text, image, audio, and video modalities. Each modality exhibits distinct semantic features and distribution patterns, covering various latent spaces. This highlights that all-modality alignment cannot rely solely on data from specific modalities; rather, it requires the integration of data across modalities.}
    \label{fig:sub_distribution}
    \vspace{-1.0em}
\end{figure*}

On the other hand, Reinforcement learning from human feedback (RLHF) plays a significant role in aligning models with human intentions \citep{ouyang2022training}. GPT-4 \citep{achiam2023gpt} has effectively boosted the model's instruction-following using RLHF. LLaMA series models \citep{touvron2023llama} have significantly improved the performance in code, mathematics, and reasoning through the post-training method, \textit{e.g.}, DPO \citep{rafailov2024direct}. 
However, this line of work is limited to single modality.
Researchers try to apply RLHF or DPO directly to multimodal scenarios, such as models with text and image modalities \citep{sun2023aligning, yu2024rlhf}. However, these methods, which rely on aligning responses with binary human preferences (\textit{e.g.}, one answer is better than another) \citep{ziegler2019fine}, struggle to be effective with more complex and diverse modalities. ImageBind \citep{girdhar2023imagebind} propose using single modality information as a \textit{binding}, such as using images as the bridging feature to learn joint embeddings. 
\emph{So, how can we establish a unified preference modeling to ensure effectiveness across any modalities?}

All-modality models refer to \textbf{models capable of accepting input and generating output in any modality}, essentially functioning as any-to-any models. Language can serve not only as a binding across different modalities but also as a natural human preference carrier. In this work, we propose utilizing language feedback to unify human feedback across all modalities, marking the first attempt to extend RLHF/DPO to arbitrary modality space and promoting the development of a general all-modality model. Overall, our work makes the following contributions:

\begin{itemize}[left=0.1cm]
    \item \textbf{Data} (\cref{sec:datasets}): The first all-modality human preference dataset -- \textit{align-anything-200k} -- on text, image, audio, and video modalities. Utilizing a two-stage human annotation process, this dataset accurately captures genuine human preferences, aiming to improve the alignment of all-modality models with human intentions.
    \item \textbf{Algorithm} (\cref{sec:algorithm}): The first algorithm applicable to enhance RLHF/DPO across all modalities: \textit{learning from language feedback (LLF)}. Our empirical results demonstrate that LLF enhance RLHF performance across 5 modalities, 5 open-source models, and 7 popular benchmarks, which achieves an average 5.83 times improvement than the baseline RLHF. 
    \item \textbf{Evaluation} (\cref{sec:evaluation}): To our knowledge, there is no system specifically designed for all-modality model evaluation. To address this gap, we propose the first evaluation tool (\textit{eval-anything}), tailored for all-modality evaluation. This tool not only meticulously constructs evaluation tasks across various modalities but also emphasizes assessing the unique features of all-modality models, such as modality selection and synergy.
    \item \textbf{Framework and Open Source}: To further the alignment of the all-modality model with human intentions, we have released all of our resources to the public.
\end{itemize}

\begin{figure}[h]
    \centering
    \includegraphics[width=\columnwidth]{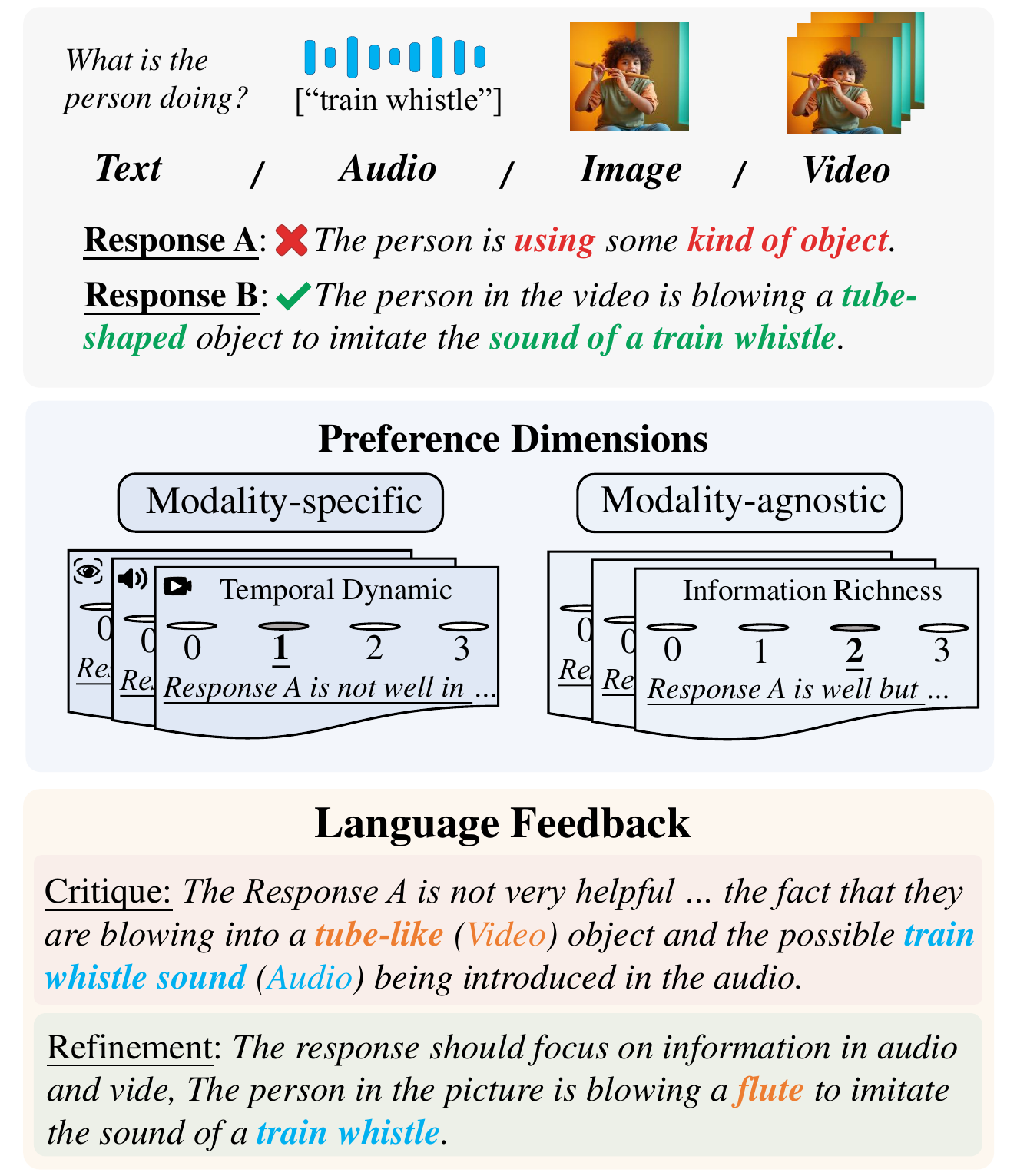}
    \vspace{-2.0em}
    \caption{\textbf{All-modality preference and language feedback annotation of \textit{align-anything-200k}.}
    For all-modality preference annotation, we classify the instruction-following metrics into two categories: \emph{modality-agnostic} and \emph{modality-specific}. Each fine-grained dimension is assigned a corresponding score along with a rationale. Additionally, we offer detailed language feedback, including critiques and refinement suggestions, which integrate information from multiple modalities within the responses.
    }
    \label{fig:dataset_language_feedback}
    \vspace{-1.0em}
\end{figure}

\section{Datasets}
\label{sec:datasets}
A primary challenge lies in the fact that existing preference datasets are predominantly focused on single-modal tasks \citep{cui2023ultrafeedback, yu2024rlhf, kirstain2023pick, majumder2024tango}, lacking comprehensive datasets that encompass all modalities, thus limiting further research progress.

In response, we open-source the first all-modality human preference dataset -- \emph{align-anything-200k} -- to enhance the instruction-following capabilities of all-modality models across tasks such as question answering, complex reasoning, \textit{etc}.
During the annotation process, we observed that introducing additional modal information significantly reduced binary human preference consistency (as shown in \cref{tab:human_preference_agreement}). Therefore, in addition to annotating fine-grained binary preferences, we introduce language feedback as an additional constraint to more accurately capture human preferences. In short, our dataset includes the following:

\begin{itemize}[left=0.2cm]
    \item \textbf{All-modality Human Preference Dataset.} We annotate human preference data across various modalities (including text, image, video, and audio) with a focus on instruction-following capabilities (\cref{fig:sub_distribution}).
    \item \textbf{One More Thing -- Language Feedback.} In addition to the widely used binary preferences, we incorporate language feedback (\emph{critique} and \emph{refinement}) to improve human preference consistency across all modalities.
\end{itemize}

\subsection{All-Modality Human Preference Dataset}
\label{subsec: all_modality_preference}
The \textit{align-anything-200k} aims to capture human preferences across 8 subtasks of all modalities, as shown in \cref{fig:sub_distribution}. With the increase in modalities, the complexity and inconsistency of human intent preferences rise significantly, driven by the unique semantic characteristics of each modality.
To achieve consistent all-modality preference modeling, we decouple annotation targets into two types, which serve as evaluation metrics for instruction-following:

\begin{itemize}
    \item \textbf{Modality-agnostic} refers to dimensions that are applied universally across modalities, including: (1) \textit{Prompt adherence}, which requires responses to be consistent with the input prompts, accurately reflecting the specified elements. (2) \textit{Rule conformity}, where responses adhere to logical, physical, or scientific principles related to the scenario or theme described in the prompt. (3) \textit{Information richness}, which emphasizes that responses should be thorough and detailed in addressing the query.
    \item \textbf{Modality-specific} represents preference dimensions tailored to the characteristics of each modality. For example, in video-related subtasks, we introduce three additional dimensions, \textit{temporal consistency}, \textit{content coherence} and \textit{motion naturalness} to better evaluate the details in the output regarding the duration, dynamics, \textit{etc.}
\end{itemize}

By decomposing the instruction-following dimensions, we establish the evaluation standards for all-modality preference annotation. More details about the annotation document can be found in Appendix: Datasets.
\begin{table*}[ht]
\centering
\footnotesize
\begin{tabular}{c>{\columncolor[gray]{0.92}}cccccccc}
\toprule
 & \textbf{T2T} & \textbf{TI2T} & \textbf{T2I} & \textbf{TI2TI} & \textbf{TV2T} & \textbf{T2V} & \textbf{TA2T} & \textbf{T2A} \\ \midrule
\textbf{w/o Language Feedback (\%)} & 73.2 ± 2.1 & 62.5 ± 7.3 & 63.2 ± 6.3 & 62.7 ± 7.7 & 62.1 ± 5.7 & 60.8 ± 4.6 & 61.8 ± 6.9 & 58.7 ± 5.8 \\ \midrule
\textbf{w/ Language Feedback (\%)} & \textbf{74.1 ± 1.2} & \textbf{70.5 ± 3.4} & \textbf{69.2 ± 2.2} & \textbf{68.3 ± 2.1} & \textbf{67.2 ± 1.8} & \textbf{65.8 ± 1.2} & \textbf{68.6 ± 2.4} & \textbf{64.3 ± 1.7} \\ \bottomrule
\end{tabular}
\vspace{-0.5em}
\caption{\textbf{Human preference agreement across different modalities.}
The results demonstrate that traditional annotation struggles to capture modality-specific details, whereas language feedback enhances consistency by conveying fine-grained human preferences effectively.}
\label{tab:human_preference_agreement}
\vspace{-1.0em}
\end{table*}

\subsection{One More Thing -- Language Feedback}
During the annotation process, there exists a decline in human preference consistency when directly conducting binary preference annotations. As illustrated in \cref{tab:human_preference_agreement}, the introduction of different modalities makes it challenging for binary preferences to fully capture human preference \citep{ji2023ai, liang2024rich, scheurer2023training}.
To address this issue, we introduce \emph{language feedback}, utilizing natural language to describe discrepancies between the model's response and the standard criteria, thereby improving the consistency of human annotation.

Specifically, we provide \emph{language feedback} for each response which includes two parts: \emph{critique}, which assesses the strengths and weaknesses of the response based on detailed criteria, and \emph{refinement}, which offers specific improvement suggestions to enhance the models' instruction-following capabilities. Through the quality control process with annotators and training verification (\cref{sec:algorithm}), we identify two key advantages of using language feedback: (1) as a natural carrier of human preferences, it clearly identifies fine-grained defects and areas for improvement in the responses, providing clearer guidance compared to binary preferences \citep{bai2022constitutional}. (2) as a connecting binding, it bridges human intent across different modalities and creates a unified, modality-agnostic preference modeling approach, capable of handling more modalities than typical methods. Further examples can be referred to in Appendix: Datasets.

\subsection{Dataset Construction}
We collect initial prompts from 24 multimodal datasets and refine them based on modality-agnostic and modality-specific dimensions with existing multimodal models (\textit{e.g.}, GPT-4o \citep{openai2024gpt4o}). For example, we emphasize dynamic temporal aspects for video or quality characteristics for audio to better adapt to each modality.
Finally, we gather the responses from 27 models for the 8 subtasks. More details can be referred to in Appendix: Datasets.

\textbf{Annotation Process}
Inspired by PKU-SafeRLHF \citep{ji2024pku}, we utilize a Human-AI joint annotation process to conduct preference annotations across fine-grained dimensions for each subtask.
As shown in \cref{fig:dataset_language_feedback}, for each fine-grained dimension's binary preferences, we assign scores from 0 to 3 following strict guidelines and detailed rationales.
Additionally, we collect language feedback from human annotators and API-based models. The annotators provide language feedback for each response by defining the critique scope, performing the critique, offering refinement suggestions, and organizing the critique and refinements into comprehensive language feedback.
As shown in \cref{fig:dataset_language_feedback}, language feedback provides more fine-grained improvement directions for cross-modal responses, serving as a carrier for expressing natural human preferences across modalities.

\textbf{Human and AI Agreement Analysis} 
We integrate API-based multimodal models (\textit{e.g.}, GPT-4o, and others) into human evaluation for binary preferences annotations. 
To analyze the consistency of the dataset with human preferences, we sample 100 data pairs from each subtask, with 10 annotators to pair-wise comparisons, and statistically measure the preference consistency, as shown in \cref{tab:human_preference_agreement}.
The results demonstrate that with the traditional binary preferences annotation pipeline, agreement consistency tends to decline as multimodal information is incorporated. This suggests that the conventional pipeline struggles to scale effectively across all modalities. As language-based feedback offers targeted and fine-grained human preference information, higher consistency is achieved across all modalities.

\section{Learning from Language Feedback}
\label{sec:algorithm}
\begin{figure*}[t]
    \centering
    \includegraphics[width=1.0\linewidth]{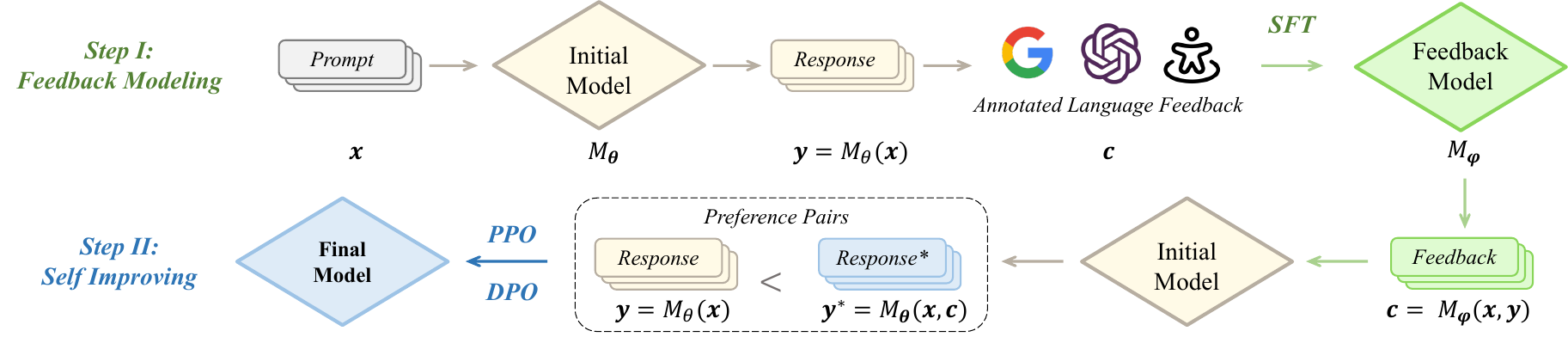}
    \vspace{-1.2em}
    \caption{\textbf{Learning from language feedback pipeline}: (1). \textit{Feedback Modeling.} We perform SFT on the initial model using annotated language feedback. (2). \textit{Self Improving.} The initial model optimizes responses given the language feedback to synthesize preference pairs.}
    \label{fig:algo}
\end{figure*}

In this section, we introduce \textit{learning from language feedback (LLF)}. It utilizes language feedback to optimize responses, synthesizing preference data which can enhance the performance of all-modality alignment.  Firstly, we review the RLHF pipeline including PPO \citep{ouyang2022training} and DPO \citep{rafailov2024direct}, highlighting the limitations of binary preferences feedback. Then we demonstrate how to practically implement LLF, including two main stages, \textit{feedback modeling} and \textit{self improving.} Finally, we empirically verify that LLF achieves an average 5.83 times improvement across 5 modalities, 5 open-sourced models, and 7 popular benchmarks. 

\begin{table*}[t!]
\renewcommand{\arraystretch}{1.3}
\centering
\resizebox{\textwidth}{!}{
\begin{tabular}{@{}lccccc>{\columncolor{lightgray}[0pt][\tabcolsep]}cc>{\columncolor{lightgray}[3pt][0.1\tabcolsep]}c@{}}
\toprule
\multirow{2}{*}{\textbf{Modality}} & \multirow{2}{*}{\textbf{Dataset}} & \multirow{2}{*}{\textbf{Focus}} & \multirow{2}{*}{\textbf{Initial Model}} & \multicolumn{5}{c}{\textbf{Performance}} \\
 &  &  &  & Initial Model & DPO & \cellcolor{white}DPO + LLF & PPO & \cellcolor{white}PPO + LLF \\ \midrule
\multirow{4}{*}{TI2T} & \multirow{2}{*}{LLaVA-Bench\cite{liu2024visual}} & \multirow{2}{*}{Visual QA} & LLaVA-1.5-7B & 90.03 & \textcolor{deepred}{88.20 (-1.83)} & 97.36 (+7.33) & 92.67 (+2.64) & \textbf{98.76 (+8.73)} \\
 &  &  & LLaVA-1.5-13B & 90.46 & 98.36 (+7.90) & 100.33 (+9.87) & 95.63 (+5.17) & \textbf{100.59 (+10.13)} \\ 
 & \multirow{2}{*}{MIA-Bench\cite{qian2024mia}} & \multirow{2}{*}{Layered Visual QA} & LLaVA-1.5-7B & 61.15 & 64.30 (+3.15) & 65.32 (+4.17)  & 70.94 (+9.79) & \textbf{72.59 (+11.44)} \\
 &  &  & LLaVA-1.5-13B & 70.34 & 70.72 (+0.38) & 75.43 (+5.09) & 76.57 (+6.23) & \textbf{80.26 (+9.92)} \\ \cmidrule(l){2-9} 
\multirow{3}{*}{T2I} & ImageReward\cite{xu2024imagereward} & Fine-Grained Preference & Chameleon-7B & -0.80 & -0.65 (+0.15) & -0.46 (+0.34) & -0.52 (+0.28) & \textbf{-0.44 (+0.36)} \\
 & HPS v2\cite{wu2023human} & Coarse-Grained Preference & Chameleon-7B & 25.19 & 25.56 (+0.37) & 25.62 (+0.43) & 25.56 (+0.37) & \textbf{25.71 (+0.52)} \\
 \cmidrule(l){2-9} 
\multirow{1}{*}{TI2TI} & InterleavedBench\cite{liu2024holistic} & Text-Image Interleaved QA &Chameleon-7B & 1.63 & 1.65 (+0.02) & 1.70 (+0.07) & \textcolor{deepred}{1.35 (-0.28)} & \textbf{1.78 (+0.15)} \\ 
\cmidrule(l){2-9} 
TV2T & Video-ChatGPT\cite{Maaz2023VideoChatGPT} & Detailed Video QA & Qwen2-VL-7B & 3.01 & 3.03 (+0.02) & \textbf{3.29 (+0.28)} & 3.02 (+0.01) & 3.05 (+0.04) \\ \cmidrule(l){2-9} 
TA2T & AIR-Bench\cite{yang2024air} & Mixed-Audio QA & Qwen2-Audio-7B & 6.61 & 6.64 (+0.03) & \textbf{6.71 (+0.10)} & \textcolor{deepred}{6.49 (-0.12)} & 6.63 (+0.02) \\ \bottomrule
\end{tabular}
}
\vspace{-0.5em}
\caption{\textbf{Main experiment results.} We validate that LLF can enhance the performance of RLHF across 5 modalities, 5 open-source models, and 7 popular benchmarks, which achieves an average 5.83 times improvement than the baseline RLHF.}
\label{tab:main_results}
\end{table*}

\subsection{Background and Preliminary}
\textbf{PPO} consists of two main stages including: \textit{step 1: preference modeling} and \textit{step 2: policy optimization}. The former involves the collection of comparison data, essential for training the reward model $r_{\text{RM}}(\cdot|\cdot)$. The process starts with response pairs $(\vy_1, \vy_2)$ generated by the initial model from shared prompts $\vx$. Human annotators are then tasked with selecting their preferred response from each pair, denoted as $\vy_w\succ\vy_l\mid\vx$, where $\vy_w$ and $\vy_l$ denote the preferred and dispreferred answer. The latter (step 2) is guided by the $r_{\mathrm{RM}}(\cdot|\cdot)$. This process is commonly modeled as a bandit setting, where a reward is obtained from the $r_{\mathrm{RM}}$ at the end of each response. 
The RL objective is,
\begin{align*}
\theta_{\mathrm{RL}} =
\mathop{\arg\max}\limits_{\theta} ~ \mathrm{E}_{\vx\sim\mathcal{P}_\gX,\vy\sim \pi_{\theta}(\cdot|\vx)}\left[r_{\mathrm{RM}}\left(\vy\mid\vx\right)\right].
\end{align*}

\noindent\textbf{DPO} directly fine-tunes the model aligning with preference pairs $\vy_w\succ\vy_l\mid\vx$. It consolidates the two stages of \textit{preference modeling} and \textit{policy optimization} in RLHF into one stage. Let  $\theta_0$ denote the initial model parameters. The optimization objective can be rewritten as,
\begin{align*}
\theta_{\mathrm{RL}} = 
\mathop{\arg\max}\limits_{\theta} ~[\log \sigma (\beta \log \frac{\pi_{\theta}(\vy_w|\vx)}{\pi_{\theta_0}(\vy_w|\vx)} - \beta \log \frac{\pi_{\theta}(\vy_l|\vx)}{\pi_{\theta_0}(\vy_l|\vx)})].
\end{align*}

Since both PPO and DPO are modality-agnostic, intuitively, all-modality alignment can be achieved with preference pairs $\vy_w\succ\vy_l\mid\vx$, and recent works have also validated its effectiveness \cite{sun2023aligning, Qwen2-Audio, ahn2024tuning, zhou2024lima}. However, the preference of coupled responses are trade-off by different dimensions, \textit{e.g.,} their style and correctness. As the number of modalities increases, these dimensions become more complex, making it harder to figure out misaligned behavior with binary preferences \cite{liang2024rich, huh2024platonic}.

\subsection{Practical Implementation}
Inspired by Constitutional AI \citep{bai2022constitutional}, LLF comprises two steps: \textit{feedback modeling} and \textit{self-improving} as shown in \cref{fig:algo}. The former employs SFT to train the feedback model, enabling it to provide language feedback based on $\vx$ and $\vy$. The latter allows the model to refine its responses based on the language feedback $\vc$. 

\textbf{Feedback Modeling} 
The training process utilizes a dataset \( \mathcal{D} = \{(\vx_i, \vy_i, \vc_i)\}_{i=1}^{N} \), where \( N \) is the size of dataset, $ \vx_i $ denotes the prompt, $ \vy_i $ represents the response, and $ \vc_i $ is the corresponding language feedback. Let $P_{\bm{\varphi}}(\vc_i \mid \vx_i, \vy_i)$ denote the probability of the target sequence $\vc_i$ given the input sequence $(\vx_i, \vy_i)$ and the model parameters $\bm{\varphi}$, the training objective of the feedback model can be expressed by the cross-entropy loss:
\begin{align}
    \mathcal{L}_{\bm{\varphi}} = - \mathbb{E}_{(\vx_i, \vy_i, \vc_i) \sim \mathcal{D}}[ \log P_{\bm{\varphi}}(\vc_i\mid \vx_i, \vy_i)].
\end{align}

\textbf{Self Improving} 
We first collect the initial model's response $\vy$ to the given prompt $\vx$ online. Then, we gather feedback $\vc=M_{\phi}(\vx, \vy)$ from the feedback model $M_{\phi}$. Finally, we have the initial model generate responses $\vy^{*}=M_{\theta}(\vx,\vc)$ conditioned on this feedback. For example, $\vy$ may suffer from redundancy and hallucination, while $\vc$ will remind the generation of $\vy^{*}$ to avoid these problems. Based on online sampling, it often enables $\vy^{*}$ and $\vy$ to exhibit significant differences in certain aspects (\textit{e.g.,} reducing redundancy), thereby generating more learnable preference pairs.

\begin{figure*}[t]
    \centering
    \captionsetup{type=figure}
    \includegraphics[width=\textwidth]{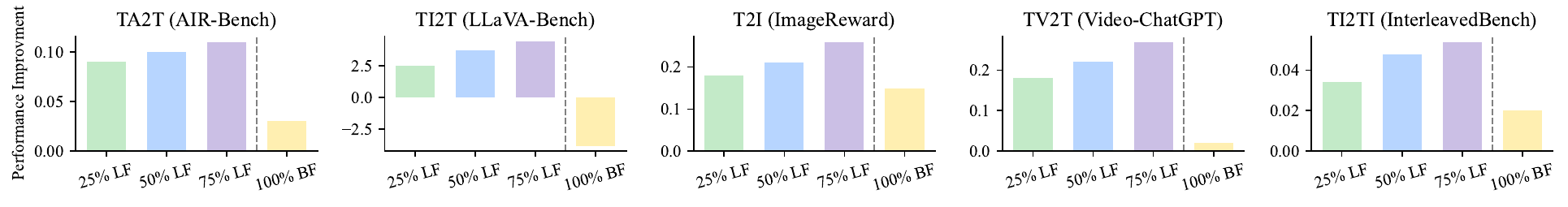}
    \vspace{-2.0em}
    \caption{\textbf{Comparison of DPO+LLF with DPO on varying language feedback amounts.} We trained the feedback models using 25\%, 50\%, and 75\% of the language feedback (LF) compared to binary feedback (BF), then synthesized an equal amount of preference pairs based on them, and subsequently compared the performance of the DPO against the initial model. We find that a small amount of language feedback can synthesize preference pairs that surpass those derived from binary feedback.}
    \label{fig:exp_2}
\end{figure*}

\subsection{Experiment} 
We empirically verify that LLF offers several key advantages over traditional binary feedback: \textit{unified preference}: As language feedback typically optimizes responses along key dimensions, models can easily learn genuine human preference beyond pairs; \textit{rich information}: Since the feedback model can generate a substantial amount of language feedback, LLF can effectively serve as a robust synthesizer of preference data. Specifically, we train Chameleon-7B \citep{team2024chameleon} for T2I and TI2TI, LLaVA-7B and LLaVA-13B \citep{liu2024visual} for TI2T, Qwen2-Audio-7B \citep{chu2024qwen2} for TA2T, and Qwen2-VL-7B \citep{Qwen2VL} for TV2T based on \textit{align-anything-200k}.
For more details about the training setting, please refer to the Appendix: Training Details.

\textbf{Improvement with Unified Preference} 
As shown in \cref{tab:main_results}, LLF synthesized preference pairs reflect more unified human preference, enhancing all-modality alignment performance. We observe that DPO and RLHF using binary pairs fall short in some modalities. However, with LLF, they yields positive improvement across all modalities. Interestingly, we find that the improvement of LLF on LLaVA-13B is greater than that on LLaVA-7B. This suggests that LLF performs better on stronger models.

\textbf{Efficiency with Rich Information}
LLF provides richer information and supports efficient preference data synthesis. As shown in \cref{fig:exp_2}, despite having only a smaller amount of language feedback, DPO+LLF outperforms DPO with binary feedback. At the same time, the reduction in data amount does not significantly weaken the capabilities of LLF. This suggests that in all-modality alignment, labeling many binary preference pairs is less effective than labeling a smaller amount of language feedback.

\begin{table}[t]
\renewcommand{\arraystretch}{1.2}
\centering
\resizebox{0.45\textwidth}{!}{
\begin{tabular}{@{}lccccc@{}}
\toprule
Modality & TI2T & TA2T & TV2T & T2I & TI2TI \\ \midrule
w/o feedback model & 44.8\% & 47.5\% & 39.6\% & 33.4\% & 41.8\% \\
w/ feedback model & 57.8\% & 64.5\% & 63.8\% & 56.9\% & 67.2\% \\ \bottomrule
\end{tabular}
}
\vspace{-0.5em}
\caption{\textbf{Abalation study of feedback model.} We compare the win rate of self-improving models against initial models on all-modality hold-out evaluation prompts. The \textit{w/o feedback model} refers to using the initial model as the feedback model, judged by GPT-4o (TI2T, T2T, TI2TI) and Gemini 1.5 Pro (TA2T, TV2T).
\vspace{-0.9em}
}
\label{tab:ablation}
\end{table}

\textbf{Ablation Study: Feedback Model is Necessary} 
As shown in \cref{tab:ablation}, the feedback model generates language feedback on the initial model’s responses to hold-out evaluation prompts, which is then used to regenerate these responses. Results indicate that this approach yields improvement while replacing the feedback model with the initial model does not achieve comparable positive effects.

\section{Evaluation: \textit{Eval-Anything}}
\label{sec:evaluation}
Currently, evaluating all-modality models relies on human experts for assessments, which is inefficient and costly. 
While combining benchmarks for individual modalities could offer a broader evaluation \citep{zhang2024lmms}, differences in data preparation, post-processing, and metrics across benchmarks hinder accurate performance assessment. 
Additionally, all-modality models uniquely select the appropriate modalities based on user queries, enabling seamless cross-modal synergy, a capability that traditional single-modality evaluation pipelines fail to capture fully.

To address this gap, we deliver our evaluation framework specifically designed for all-modality models -- \textit{eval-anything} -- including (1) \textbf{all-modality understanding} (AMU) for assess models to simultaneously process and integrate information from all modalities and (2) \textbf{all-modality generation} (AMG): evaluate a model’s ability to follow user instructions, autonomously select modalities, and work synergistically across different modalities for output.

\begin{figure*}[ht]
    \centering
    \includegraphics[width=1.0\linewidth]{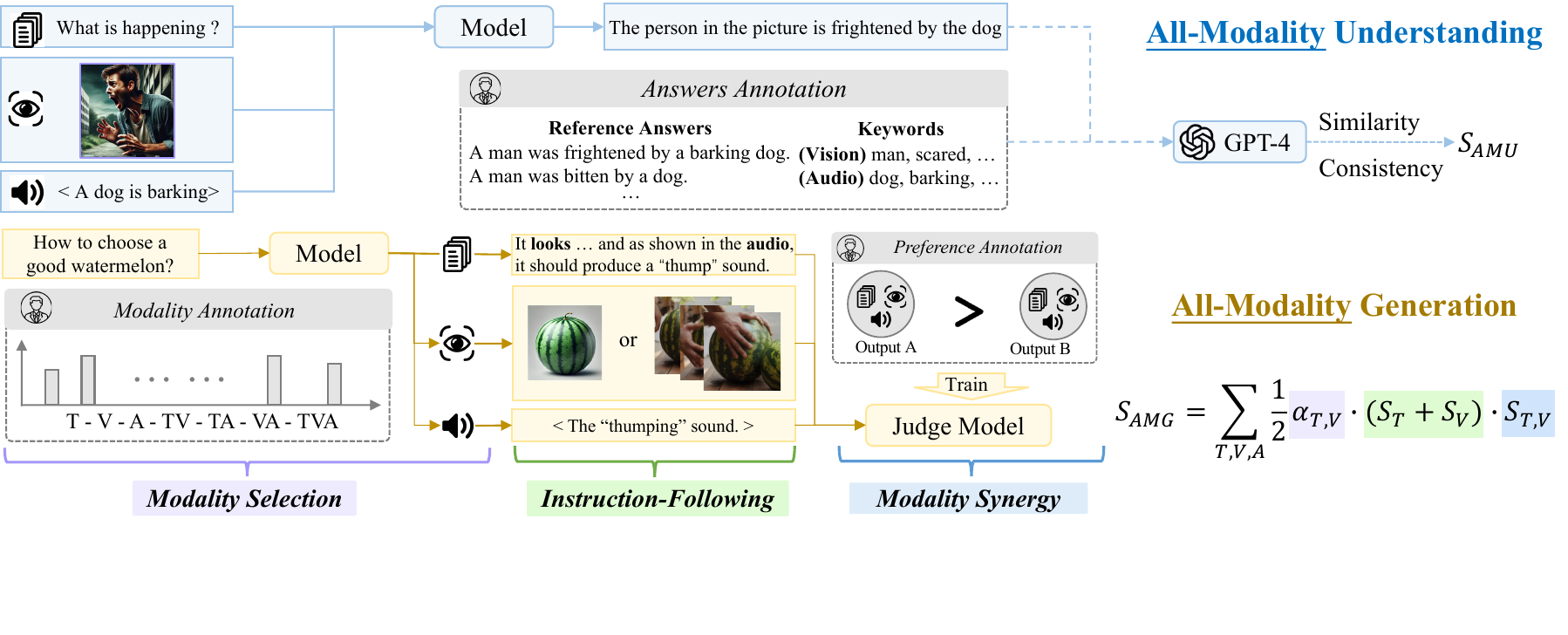}
    \vspace{-5.2em}
    \caption{\textbf{The \textit{eval-anything} benchmark consists of two components: (Up) }\textbf{\textit{{AMU: All-Modality Understanding}}}, where the model answers open-ended questions by integrating textual instructions, images, videos, and audio. 
    \textbf{(Down) \textit{AMG: All-Modality Generation}} is divided into subtasks of instruction-following, modality selection, and synergy. The model generates outputs for each modality (text, image, video, audio) based on instructions, with human-preferred combinations guiding modality selection metrics. A trained judge model evaluates the relevance, consistency, and synergy across different modalities in the outputs.
    }
    \label{fig:evaluation_pipeline}
\end{figure*}

\subsection{Composition of Evaluation Dataset}

\subsubsection{All-Modality Understanding}

All-modality models aim to both understand individual modalities and combine information across them to generate high-quality responses. 
To assess their comprehensive multimodal processing, we create 164 test entries, each containing textual, visual (image or video), and auditory (audio or speech) components.
These interconnected modalities require the model to integrate all inputs accurately, as failure in any one modality leads to incorrect answers.
For instance, as shown in \cref{fig:evaluation_pipeline}, if a model fails to process visual inputs, it may miss that \emph{the person in the picture is frightened}. Similarly, without auditory processing, it might not understand that the person’s fear is due to \emph{a barking dog}.

\begin{figure}[t]
    \centering
    \includegraphics[width=0.9\columnwidth]{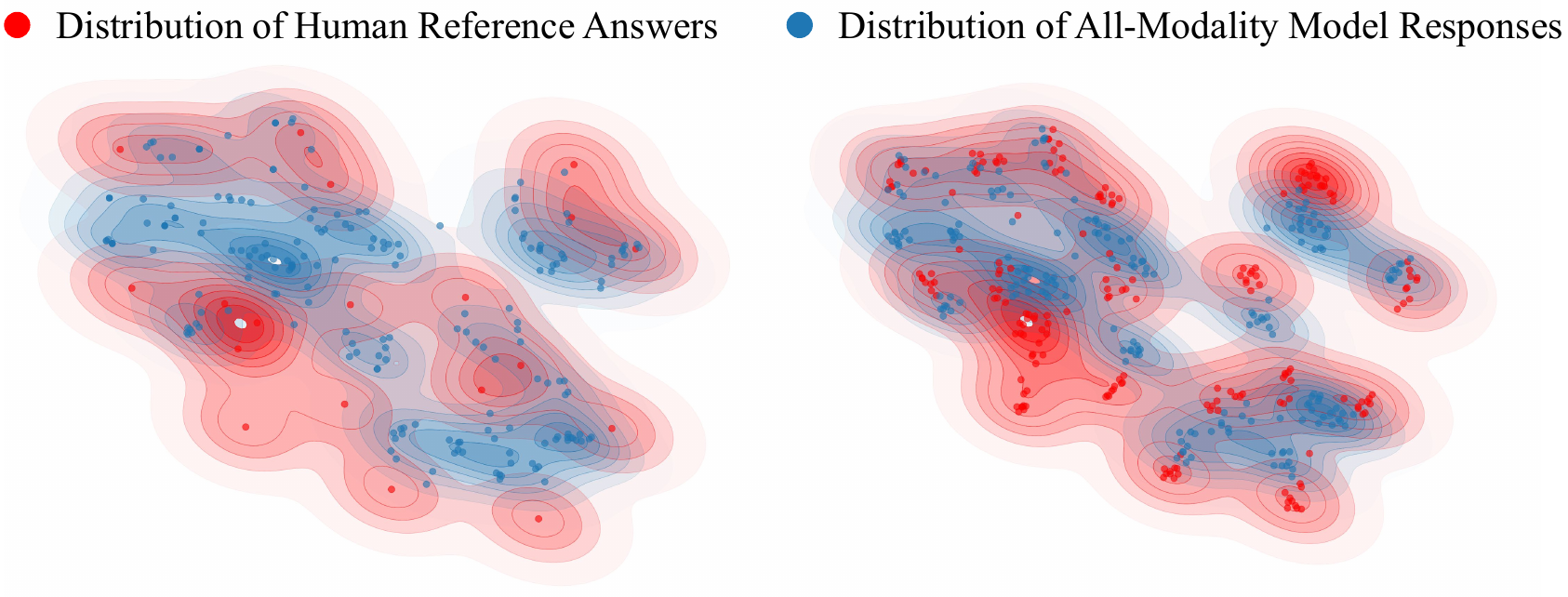}
    \vspace{-0.5em}
    \caption{\textbf{Distribution of model and human annotation.} 
    In AMU task, comparing the distribution of an all-modality model’s responses (blue) with human annotations (red) reveals that a single annotation inadequately covers the model's semantic space (left), whereas multiple annotations broaden the red region, improving evaluation coverage by capturing more response diversity (right).
    }
    \label{fig:amu_reference_visualization}
    \vspace{-0.8em}
\end{figure}
We use GPT-4 \citep{achiam2023gpt} to evaluate model responses on a scale of 1 to 10. 
However, previous studies \citep{liu2024visual, yang2024air} underscore limitations in multimodal evaluation that rely on a single human annotation as a reference. 
As the number of modalities increases, so does the complexity, making it harder to reach consensus and increasing subjective bias in single annotations. Moreover, since GPT-4 lacks true multimodal comprehension, evaluating only its text responses doesn’t confirm if essential information from each modality is fully understood. To address this, we gather responses from 10 annotators and extract key terms from each modality.
As illustrated in \cref{fig:amu_reference_visualization}, the distribution of multiple annotations mitigates the bias associated with single-reference evaluation. 
Additionally, the inclusion of key terms enables GPT-4 to more accurately detect potential errors in its responses.
The details for AMU can be found in Appendix: Evaluation.

\subsubsection{All-Modality Generation}
In generation tasks, all-modality models can outperform single-modal ones by delivering diverse information across multiple formats under the same user query. To achieve this, they must follow \textit{\textbf{SSI}} principles: (1) \textbf{S}elect relevant modalities automatically to reduce redundancy, (2) \textbf{S}ynergistic integration for maximum information gain, and (3) \textbf{I}nstruction-following in each modality.
The overall score for the AMG task is as follows:
\begin{align}
    S_\text{AMG} = \sum_\textit{T, V, A} \frac{1}{2} \, \alpha_\textit{T, V} \cdot \left( S_\textit{T} + S_\textit{V} \right) \cdot S_{\textit{T, V}}\;.
\end{align}

\textbf{Instruction Following}
To assess instruction-following, we design 100 prompts per modality. Using GPT-4o, we score the text and image outputs. For audio and video modalities, inspired by TIFA \cite{hu2023tifa} pipeline, we create multiple-choice questions based on the prompts and employ Qwen2-Audio \cite{Qwen2-Audio} and Qwen2-VL \cite{Qwen2VL} as evaluation models. This process yields independent modality scores $S_\textit{T}, S_\textit{V}, S_\textit{A}$, each ranging from 0 to 10.

\textbf{Modality Selection}
Properly combining modalities in the model's response can provide rich perspectives while reducing redundant information. We employ 25 crowdsourced annotators to identify the expected modality combinations for given text instructions. The voting process results in one of the quantitative metrics for AMG task, with appropriate rewards and penalties applied based on the modalities generated by the model. As shown in \cref{fig:evaluation_pipeline}, $\alpha$ represents the percentage of human votes for each modality combination, and if the model outputs all three modalities, this parameter will be one-third of $\alpha_\textit{T, V, A}$.

\textbf{Modality Synergy}
refers to the consistency between different modalities in a model’s responses. We assess modality synergy by training a judge model with human annotations obtained through a data synthesis. Specifically, we develop a data synthesis pipeline centered on LLM-based agents that utilize tools to invoke audio, image, and video generation models, constructing a dataset with textual inputs and multimodal outputs. We employ human annotators to annotate preference data considering the synergy between different modalities in each response, and then train a judge model that allows all-modality input. As shown in \cref{fig:evaluation_pipeline}, responses with high synergy scores ($S_\textit{T, V}, S_\textit{T, A}, S_\textit{V, A}$) should show high relevance and consistency between modalities, with visual and auditory elements enriching the text from various perspectives. More details on training the judge model can be found in Appendix: Evaluation.

\begin{table*}[ht]
\centering
\renewcommand{\arraystretch}{1.2}
\resizebox{1.0\textwidth}{!}{
\begin{tabular}{lcccccccccccccc}
    \toprule
     \multirow{3}{*}{\textbf{Initial Models}} & \multicolumn{5}{c}{\textbf{All-Modality Understanding}} & \multicolumn{8}{c}{\textbf{All-Modality Generation}}& \multirow{3}{*}{\textbf{Overall}}\\
     \cline{2-6}
     \cline{7-14}
     ~ & \multicolumn{4}{c}{\textbf{Category}} & \multirow{2}{*}{\parbox{0.8cm}{\centering \textbf{AMU} \textbf{Score}}} & \multirow{2}{*}{\parbox{1.2cm}{\centering \textbf{Modality} \textbf{Select}}} & \multicolumn{3}{c}{\textbf{Instruction Following}}& \multicolumn{3}{c}{\textbf{Modality Synergy}} & \multirow{2}{*}{\parbox{0.8cm}{\centering \textbf{AMG} \textbf{Score}}} & \\
     \cline{2-5}
     \cline{8-10}
     \cline{11-13}
     ~ & Perception & Reasoning & IF & Safety &  &  & T & V & A & T-V & T-A & V-A &  & \\
     \midrule
     LLaVA-v1.5-7B$^\dag$ & 2.66 &2.67 & 2.50 &2.90 &2.68 & 0.182 & 5.62 & 9.47 & 4.73 & 0.29 & 0.33 & 0.66 & 1.56 & 2.12 \\
     Qwen2-VL$^\dag$ & 2.76 & 3.07 &2.40 &4.05 &3.07 & 0.177 & 6.70 & 9.51 & 4.83 & 0.52 & 0.57 & 0.66 & 2.16 & 2.62 \\
     Qwen2-Audio$^\dag$ & 3.58 &4.53 &3.40 &2.65 &3.54 & 0.190 & 5.93 & 9.30 & 4.85 & 0.42 & 0.46 & 0.66 & 1.97 & 2.75 \\
     Chameleon-7B$^\dag$ & 1.44 &2.97 &2.80 &2.45 &2.41 & 0.156 & 4.21 & 9.39 & 4.66 & 0.45 & 0.47 & \textbf{0.70} & 1.57 & 1.99 \\
     Llama3.1-8B-Instruct$^\dag$ & 1.05 &1.20 &1.20 &1.35 &1.20 & 0.231 & 7.69 & 9.58 & 5.15 & \textbf{0.55} & \textbf{0.59} & 0.65 & 3.08 & 2.14 \\
     Gemini-1.5-Pro$^\dag$ & \textbf{5.36} & \textbf{5.67} &\textbf{6.70} & \textbf{6.70} & \textbf{6.11} & 0.227 & 9.45 & 9.65 & \textbf{6.76} & 0.44 & 0.47 & 0.66 & 3.05 & \textbf{4.58} \\
     GPT-4o$^\dag$ & 2.66 &3.48 &4.20 &5.15 &3.87 & \textbf{0.266} & \textbf{9.51} & \textbf{9.70} & 6.64 & 0.51 & 0.56 & 0.67 & \textbf{3.96} & 3.92 \\
    \bottomrule
\end{tabular}
}
\vspace{-0.5em}
\caption{\textbf{The performance of models in the eval-anything benchmark.} Additional input modalities are masked for models that do not support all-modality input. Since most current open-source models lack support for all-modality output, (\textbf{$\dag$}) indicates that models are used as agents to invoke AudioLDM2-Large \cite{liu2024audioldm} and FLUX.1-schnell\cite{black2024flux} for audio and image generation.
}
\label{tab:eval_main_results}
\vspace{-1.0em}
\end{table*}

\begin{figure}[t]
    \includegraphics[width=\columnwidth]{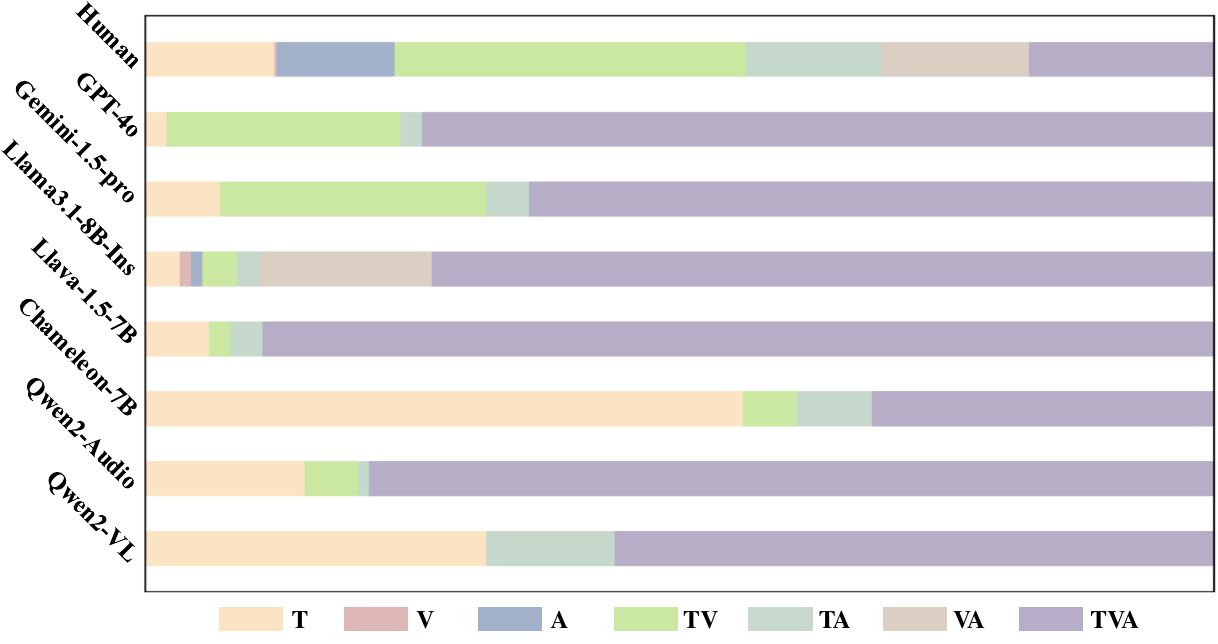}
    \vspace{-1.8em}
    \caption{\textbf{Performance of models in modality selection metric.} We analyze the model’s modality selection performance across all instructions in AMG task, comparing it to the voting results on preferred modality combinations from 25 human annotators.}
    \label{fig:modality_selection_comparison}
    \vspace{-0.8em}
\end{figure}

\subsection{Evaluation Results \& Analysis}
\paragraph{Human and AI Agreement}
In the modality synergy task, after training on the 5k preference dataset, the experiment reveals a 66.4\% agreement rate between the judging model and human annotators. These figures are consistent with human agreement ratios reported in similar studies on modeling human preferences \citep{xu2024imagereward} in the multimodal large language models domain. 

\paragraph{Input vs Outputs}
Most models in \cref{tab:eval_main_results} support partial modality input and have baseline scores, but Gemini-1.5-Pro outperforms others due to its ability to process all three modalities.
In the AMG task, the average scores is relatively low, with no model demonstrating a clear advantage across all sub-items.
The results indicate that, compared to modality generation, models are more advanced in modality understanding, consistent with the developmental trends of all-modality models.

\paragraph{Truly All-Modality Model}
Current models still fall far behind all-modality models. For all-modality understanding, models using only a single modality score less than half the maximum points, as shown in \cref{tab:eval_main_results}. 
Even Gemini-1.5-Pro, which processes both visual and auditory inputs, fails to attain a perfect score. 
Unlike humans, who integrate information from multiple modalities, these models are limited by their inability to perceive different modalities in a fully integrated way, even though they perform nearly at the human level within individual modalities.
Moreover, models align poorly with human choices when selecting output modalities, as illustrated in \cref{fig:modality_selection_comparison}. 
Humans can adaptively choose the best combination of modalities based on the instruction. In contrast, models tend to output either only textual information, resulting in information loss, or all available modalities, causing redundancy. 
The limited multimodal capability, especially in models trained mainly on text, hampers their ability to synthesize information effectively, making it difficult to balance detail and conciseness.

\section{Conclusion}
In this work, we make the first exploration of fine-tuning all-modality models using human preference data across diverse modalities, to ensure alignment with human intentions. We have open-sourced the align-anything dataset, incorporating 200k annotated human preference data across modalities. Our proposed alignment method leverages language feedback to capture complex, modality-specific human preferences, significantly enhancing the model’s ability to follow instructions. To assess the all-modality models, we developed an evaluation benchmark: eval-anything. All data, models, and code have been made openly available.

\paragraph{Ethics Impact} 
Our data collection, approved by the Institutional Review Board, will be released under the CC BY-NC 4.0 license. The dataset integrates Q-A data from open-source and API-based models, offering the potential for developing AI aligned with human intentions. However, while it could theoretically be misused for harmful purposes, we are committed to promoting safe AI technology. 

\paragraph{Limitations and Future Work}
Despite having 45 experienced annotators, the team lacks sufficient diversity, which may limit the representativeness of human preferences. To address this, we plan to recruit a more varied group via platforms like Amazon MTurk. Additionally, the absence of all-modality foundational models in open-source restricts validation to individual modalities, so we encourage community efforts to develop such models and aim to expand experiments across more models. This is an ongoing effort, with the goal of scaling the dataset to millions.

{
    \small
    \bibliographystyle{ieeenat_fullname}
    \bibliography{main}
}

\clearpage

\definecolor{myblue}{rgb}{0.388, 0.765, 0.925}

\setcounter{page}{1}
\maketitlesupplementary

\section{Ethic Responsibility}
\label{app:ethic_responsibility}
Our data collection has been approved by an Institutional Review Board (IRB). The IRB file contains institutional information. To maintain anonymity in the double-blind review process, we did not upload the IRB documents alongside the supplementary materials. If needed, we are willing to discuss the IRB file further with the Ethics Reviewer, provided it does not compromise the double-blind review protocol.

\section{Open-Source Assets and License}
All datasets examples, codes, and demos have been attached to our supplementary material. In \cref{app:ethic_responsibility}, we discuss potential risks and mitigation strategies related to model open-sourcing in detail. After the double-blind review process, we will actively engage with community feedback regarding the data, framework, and code and promptly address any issues related to version inconsistencies to further advance scientific research on large model alignment.

The following assets are planned for open-source release after the double-blind review process: 

\begin{itemize}
    \item Datasets
        \begin{itemize}
            \item Prompt (see dataset in supplementary material)
            \item Response
            \item Preference Data (view a subset of the samples in supplementary material)
        \end{itemize}
    \item Algorithms
        \begin{itemize}
            \item Learning from language feedback algorithms, covering text-to-text (T2T), text-image-to-text (TI2T), text-to-image (T2I), text-video-to-text (TV2T), text-audio-to-text (TA2T) modalities.
            \item Baseline alignment algorithms, covering supervised fine-tuning (SFT), reward modeling, and reinforcement learning from human feedback (RLHF), along with well-tuned hyper-parameters, in over 8 modalities.
        \end{itemize}
    \item Evaluation
    \begin{itemize}
        \item Dataset: Test entries, including text instructions and corresponding multi-modal data of AMU task and AMG task.
        \item Model: A judge model used for the modality synergy evaluation in AMG task.
        \item Codebase: The codebase of the \textit{eval-anything} benchmark. Moreover, an all-modality evaluation framework, covering 8 modalities, 30+ benchmarks, and 2 inference backends.
    \end{itemize}
\end{itemize}

To provide reviewers with detailed data references, we have included samples from each modality of the dataset, along with additional related materials. After the double-blind review process, we will release all content as open source.

\section{More Details of Related Works}
\label{app:related_work}
Building on the success of large language models (LLMs) and the latest advancements in multimodal large language models (MLLMs), there is growing anticipation for integrating multiple modalities into a single model to achieve truly all-modality capabilities, \emph{i.e.}, all-modality models. However, numerous challenges must be overcome to reach this goal. We will introduce the related works from the aspects of dataset, algorithms, and evaluation.

\paragraph{Dataset} 
Recently, several studies have proposed preference datasets for multimodal models, encompassing both understanding and generation tasks \citep{amirloo2024understanding, uehara2024understanding}. However, current preference datasets are restricted to specific modalities and tasks, and they lack comprehensive cross-modal annotations. 
In the domain of Text-Image-to-Text (TI2T) understanding, existing 
preference datasets primarily aim to reduce hallucinations and enhance helpfulness in MLLMs \citep{sun2023aligning, zhou2024aligning, yu2024rlhf, li2024vlfeedback}.
In the Text-to-Image (T2I) domain, most existing studies employ binary human ratings or preference rankings to reflect authentic human preferences \citep{xu2024imagereward, kirstain2023pick, wu2023human}. \citet{liang2024rich}, in contrast, gathers more comprehensive feedback, focusing on misaligned regions and key terms.
In the video and audio domains, aside from datasets such as LLaVA-Hound \citep{zhang2024direct}, SafeSora \citep{dai2024safesora}, and Audio-Alpaca \citep{majumder2024tango}, which provide binary human preference data, further research on preferences remains relatively scarce.

\paragraph{Algorithms} Recent all-modality alignment algorithms directly extend the RLHF pipeline to the corresponding modality, since both PPO \citep{ouyang2022training} and DPO \citep{rafailov2024direct} are modality-agnostic. RLHF-V \citep{yu2024rlhf} have used human feedback to annotate image-question-answering data, significantly reducing hallucinations of MLLMs. Similarly, Qwen2-Audio-7B-Instruct \citep{Qwen2-Audio} have attempted to apply this approach to audio language models, improving audio comprehension and question-answering tasks. Similar strategies have also been practiced in video question-answering \citep{ahn2024tuning}, image-generation \citep{blacktraining}, and video generation \citep{dai2024safesora} tasks. However, the preference of coupled responses are trade-off by different dimensions, \textit{e.g.,} their style and correctness. As the number of modalities increases, these dimensions become more complex, making it harder to figure out misaligned behavior with binary preferences \citep{liang2024rich, huh2024platonic}. Due to the increased difficulty and cost of annotating all-modality data \citep{yu2024rlaif}, there is an urgent need for an alignment algorithm that more efficiently utilizes human feedback.

\paragraph{Evaluation} Current benchmarks for all-modality models primarily focus on evaluating individual modalities. Numerous benchmarks are developed to assess the capabilities of large vision models across multiple dimensions, including image perception \citep{schwenk2022okvqa, fu2023mme, liu2023mmbench, yu2023mm}, object hallucination \citep{li2023evaluating, jing2023faithscore, chen2023mitigating}, and safety performance \citep{liu2023mm, wang2023tovilag, liu2024safety}. For the video modality, \citet{ning2023video, fu2024video, li2024mvbench} develop various benchmarks to assess large vision models. Furthermore, in the audio domain, there are dedicated evaluation datasets for large audio models \citep{yang2024air, weck2024muchomusic}. However, these studies are not specifically designed to assess their cross-modal comprehensive understanding capabilities. Existing evaluation methods are primarily based on a series of tasks, such as Audio-Visual Question Answering \citep{yun2021pano, yang2022avqa, li2022learning}, Audio-Visual Scene-aware Dialog \citep{alamri2019audio, schwartz2019simple}, and Audio-Visual Retrieval \citep{chen2023valor, chen2023vast}. Although involving multiple modalities, these tasks are not specifically designed to assess the all-modality capabilities of MLLMs.

\section{Align-Anything Framework}

\begin{figure}[ht]
    \centering
    \includegraphics[width=\columnwidth]{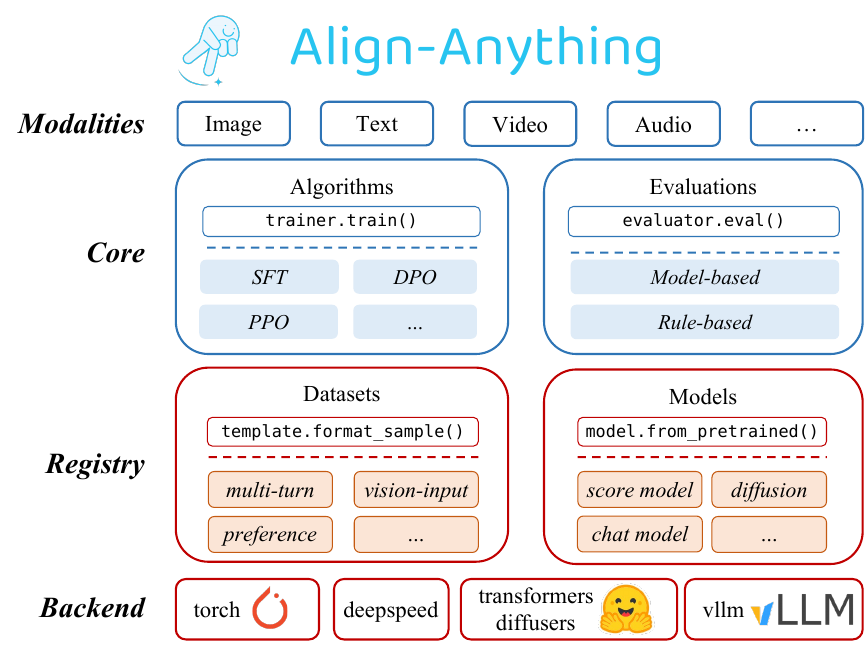}
    \vspace{-1.8em}
    \caption{\textbf{The open-sourced framework of align-anything.}}
    \label{fig:framework}
\end{figure}

\begin{figure*}[h]
    \centering
    \includegraphics[width=\textwidth]{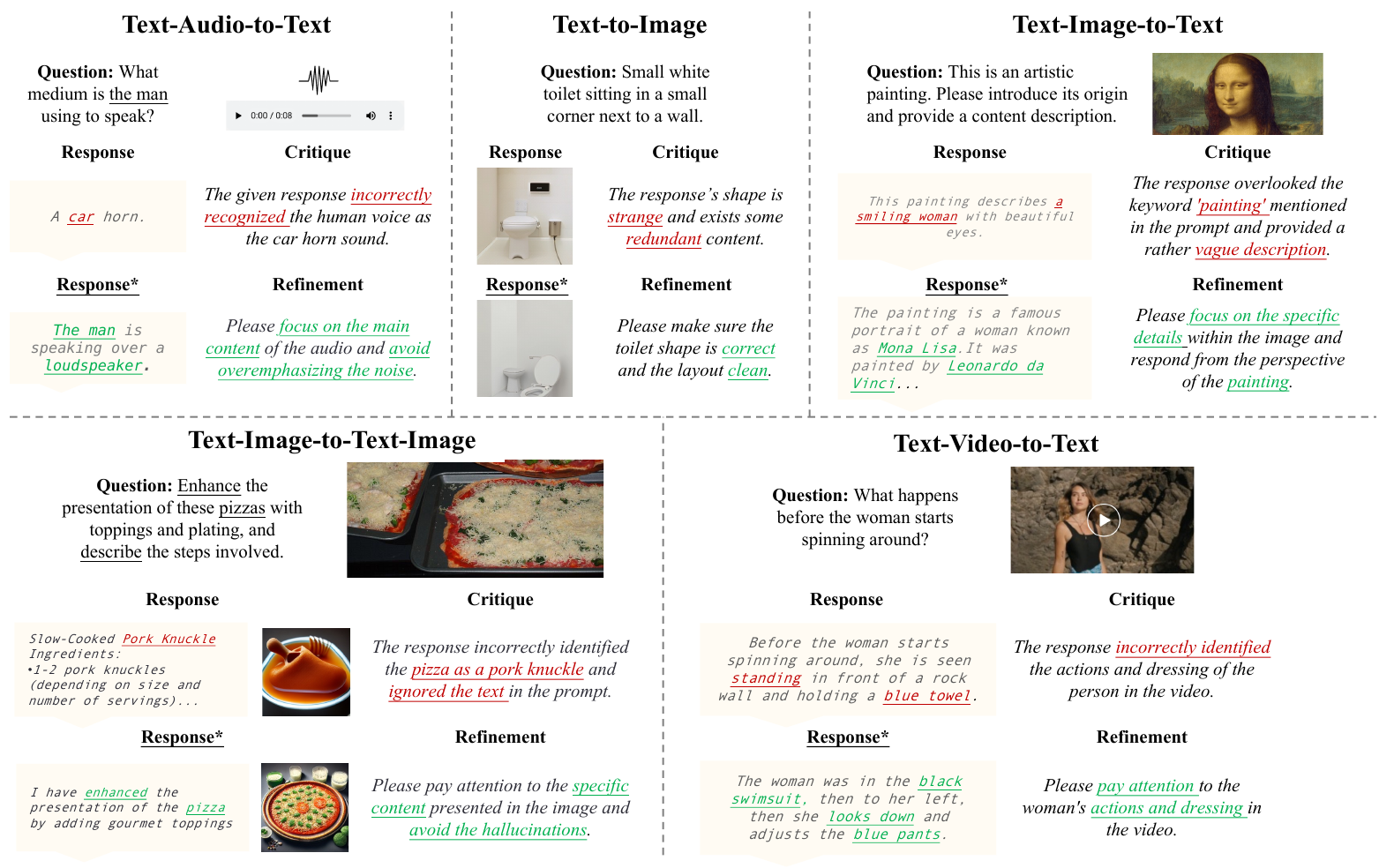}
    \vspace{-2.0em}
    \caption{\textbf{Examples of LLF synthesized preference pairs.} The responses from current models are often not perfect. Using language feedback to enhance prompts can improve responses in certain dimensions, synthesizing more learnable preference pairs.}
    \label{fig:llf_example}
\end{figure*}

\begin{figure*}[h]
    \centering
    \includegraphics[width=\textwidth]{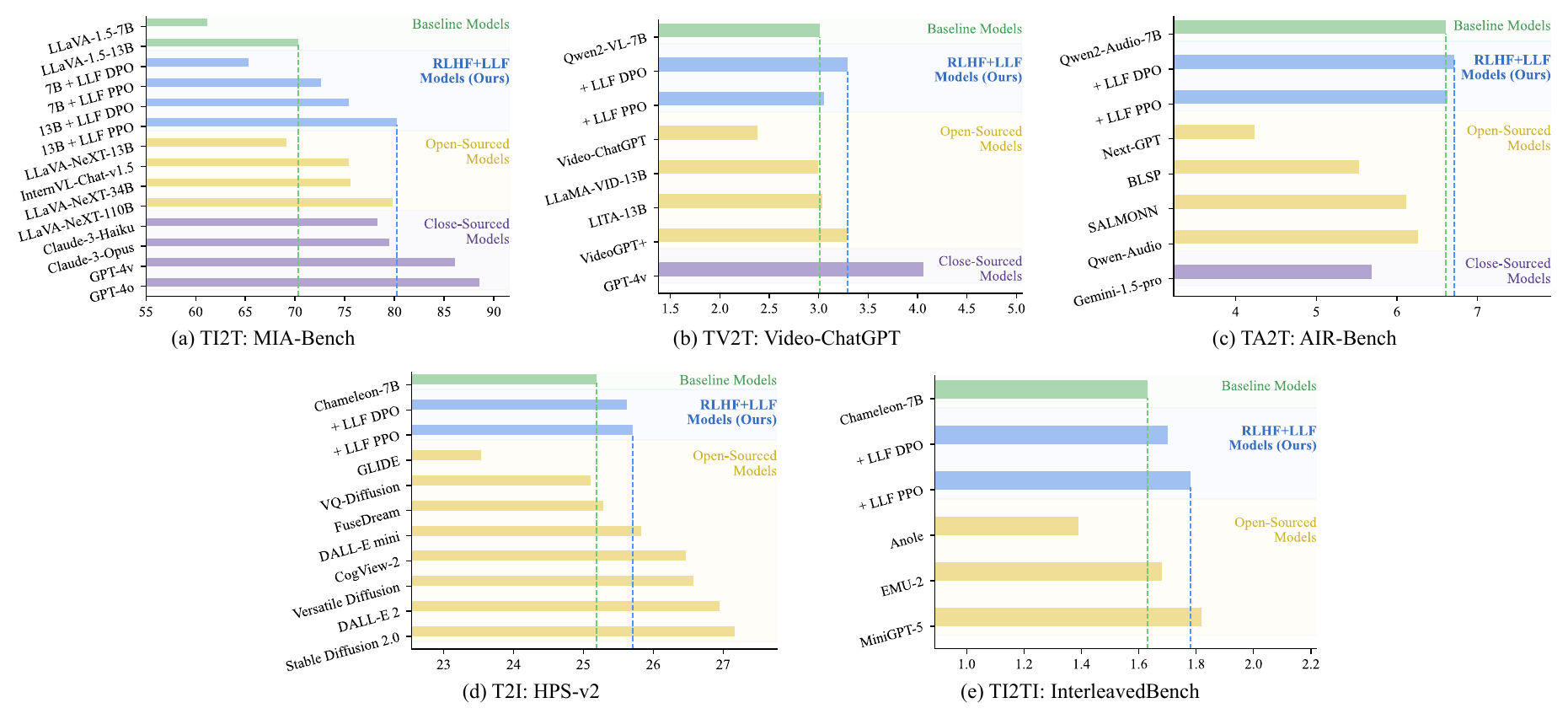}
    \vspace{-2.0em}
    \caption{\textbf{Further experiment results of LLF on TI2T, TV2T, TA2T, T2I and TI2TI modalities.} We conducted comparisons of LLF with common open-source \citep{liu2024improved, Qwen2-Audio, team2024chameleon, liu2024llavanext, qwenaudio, wu2024nextgpt, Rombach_2022_CVPR, liu2021fusedream, nichol2022glide, ding2022cogview2, chern2024anole, sun2024generative, zheng2023minigpt, wang2023blsp, tang2023salmonn} and closed-source models \citep{openai2024gpt4o, anthropic2024claude3, Maaz2023VideoChatGPT, reid2024gemini}.}
    \label{fig:llf_comparison}
\end{figure*}

Our work is built on the framework of align-anything, which is designed for training and evaluation across all modalities. As shown in \cref{fig:framework}, the align-anything framework aims to align all-modality large models, including large language models (LLMs), vision language models (VLMs), and others, with human intentions and values. Overall, this framework has the following characteristics:

\begin{itemize}
    \item \textbf{High Modularity.} Its versatility stems from the abstraction of different algorithm types and well-designed APIs, allowing users to easily modify and customize the code for different tasks.
    \item \textbf{Support for Various Model Fine-Tuning.} This framework includes fine-tuning for models such as LLaMA \citep{touvron2023llama}, LLaVA \citep{liu2024improved, liu2024visual}, Qwen2-VL \citep{Qwen2VL}, Qwen2-Audio \citep{Qwen2-Audio}\footnote{To the best of our knowledge, align-anything is the first and, until the CVPR submission deadline, the only framework that supports Qwen2-Audio RLHF fine-tuning.}, Chameleon \citep{team2024chameleon}\footnote{The same, currently, align-anything is the only framework that supports Chameleon fine-tuning and RLHF with image generation capability.}, and others.
    \item \textbf{Support Fine-Tuning across Any Modality.} It supports fine-tuning alignments for different modality models, including LLMs, VLMs, and other modalities
    \item \textbf{Support Different Alignment Methods.} The framework supports different alignment algorithms, including SFT, DPO \citep{rafailov2024direct}, PPO \citep{ouyang2022training}, and others.
\end{itemize}

\subsection{Training Part}

The align-anything framework integrates all-modality alignment algorithms (\textit{e.g.,} Reinforcement Learning from Human Feedback, RLHF), supporting SFT, RM, DPO, and PPO. Additionally, align-anything implements KTO \citep{ethayarajh2024kto}, SimPO \citep{meng2024simpo}, and ORPO \cite{hong2024reference} in the text-to-text modality. Besides, align-anything offers a highly scalable model registration mechanism and currently supports the training and deploying over 25 models. For more details, please refer to the \texttt{align-anything-code/README.md} file of our supplementary materials.

\subsection{Evaluation Part}
The align-anything evaluation framework now supports over 30 commonly used benchmarks, covering all common modalities. For more details, please refer to the \texttt{align-anything-code/README.md} file of our supplementary materials.

\section{Training Details}

This section will introduce the implementation details of \textit{learning from language feedback (LLF)}, hyper-parameter settings, case studies, and the computational devices involved in the experiments.

\subsection{Implementation Details}

LLF comprises two primary steps: \textit{feedback modeling} and \textit{self improving}. The first step employs maximum likelihood to enable the model to learn from \textit{align-anything-200k} how to generate language feedback for a given prompt and response. This includes evaluating the response (critique part) and providing suggestions for improvement (refinement part). During the self improving phase, when optimizing the response, we incorporate the refinement into the prompt to guide the model in generating better responses on specific dimensions, thereby creating preference pairs with the original response.

In the RLHF phase, we synthesized a preference dataset of the same size as the baseline data (detailed size for each modality can be found at \cref{fig:pie-chart}). We will open-source this preference dataset synthesized by LLF. \cref{fig:llf_example} provides a simple example.

\subsection{Further Comparison}

This section will compare LLF with other open-source and closed-source models. These comparison results reveal two surprising phenomena. For modalities with initially weaker models, such as TI2T, LLF can significantly enhance their performance. As shown in \cref{fig:llf_comparison}, on the MIA-Bench \citep{qian2024mia}, although the initial model LLaVA-1.5-13B \citep{liu2024improved} is not outstanding, after fine-tuning with LLF-based PPO, it surpasses open-source models with much larger parameter sizes (\textit{e.g.,} LLaVA-NeXT-110B \citep{liu2024llavanext}) and even some closed-source models (\textit{e.g.,} Claude-3-Opus \citep{anthropic2024claude3}). Meanwhile, for modalities with initially stronger models, like TA2T, LLF can still further improve their performance. 

\subsection{Models}
LLF pipeline requires multimodal models with language understanding capabilities. Considering popularity, performance, and open-source availability, we select the following models for our experiment.

\begin{itemize}
    \item \textit{LLaVA-1.5-7B \& LLaVA-1.5-13B} \citep{liu2024improved} are visual language models validated their performance across 11 benchmarks. their base language models are Vicuna-7B-v1.5 and Vicuna-13B-v1.5 \citep{peng2023instruction}, with CLIP-ViT-L-336px \citep{radford2021learning} as the visual encoder and MLP as the multimodal projection layer. The 13B version is trained on approximately 1.2M publicly available data. They excel at extracting effective information from inputs that combine language and images. Then they provide answers by integrating the world knowledge of the language model.
    \item \textit{Qwen2-Audio-7B-Instruct} \citep{Qwen2-Audio} is a large-scale audio-language model, which can process diverse audio signal inputs to perform audio analysis or generate direct textual responses to speech instructions. Qwen2-Audio-7B-Instruct uses Whisper-large-v3 \citep{radford2023robust} as the audio encoder and Qwen-7B \citep{bai2023qwen} as the language model. It was pre-trained on over 520k audio QA data and completed SFT and DPO alignment, achieving outstanding performance on 9 benchmarks.
    \item \textit{Chameleon-7B} \citep{team2024chameleon} is an early-fusion, token-based multimodal model designed for unified image and text understanding and generation. Chameleon-7B employs a newly trained image tokenizer, based on \citep{gafni2022make}, to encode $512 \times 512$ images into 1024 discrete tokens. The model was pre-trained using a Llama2-style language model architecture on text-image interleaved data with this encoder. The original paper introduced two model variants: Chameleon-7B and Chameleon-34B. Due to computational constraints, we focused on evaluating our method using Chameleon-7B. However, the full Chameleon-7B parameters with image generation capabilities were not open-sourced by the authors. To address this limitation, we used the \ours{} framework to finetune the original Chameleon-7B model on the LAION-Art dataset \citep{schuhmann2022laion}. This resulted in the AA-Chameleon-7B-base model, which integrates both image generation and understanding capabilities. For simplicity, we refer to this model as Chameleon-7B throughout our paper.
    \item \textit{Qwen2-VL-7B-Instruct} \citep{Qwen2VL} is a large multimodal model that employs a unified paradigm for processing both images and videos, representing a significant advancement in video understanding capabilities. Through its innovative approach to visual processing, the model can effectively analyze video content using the same architecture that handles static images, creating a seamless integration of temporal and spatial information for comprehensive video comprehension. Qwen2-VL-7B-Instruct uses Vision Transformer \cite{dosovitskiy2021imageworth16x16words} as the vision encoder and Qwen2-7B \cite{yang2024qwen2} as the language model. Trained on 1.4 trillion tokens and fine-tuned through SFT, the model demonstrates exceptional video comprehension capabilities.
\end{itemize}

\subsection{Benchmarks}

\begin{itemize}
    \item \textit{LLaVA-Bench (COCO)} \citep{liu2024visual} is a benchmark designed to evaluate the alignment behavior and capabilities of models with consistent visual inputs. It utilizes a subset of the COCO-Val-2014 dataset, selecting 30 images and generating three distinct types of questions for each image: conversation, detailed description, and complex reasoning. This results in a total of 90 questions, crafted using a specialized data generation pipeline. The benchmark aims to assess how effectively models can follow user instructions and handle diverse question types.
    \item \textit{MIA-Bench} \citep{qian2024mia} is a pioneering benchmark designed to assess the capabilities of VLMs in adhering to complex, layered instructions. This benchmark features a diverse collection of 400 meticulously crafted image-prompt pairs, each intended to rigorously test the models' ability to generate responses that precisely follow intricate directives. Through comprehensive evaluations of various state-of-the-art VLMs, MIA-Bench uncovers significant disparities in performance, thereby identifying key areas for enhancement in instruction fidelity.
    \item \textit{ImageReward} \citep{xu2024imagereward} is a general-purpose T2I human preference reward benchmark. Trained on 137k expert-annotated comparisons, the ImageReward model utilizes a systematic annotation pipeline emphasizing alignment, fidelity, and harmlessness in generated images. It significantly outperforms prior metrics like CLIP and FID in capturing human preferences, achieving higher consistency with human rankings. The benchmark demonstrates robustness across various datasets, establishing itself as a promising metric for evaluating T2I generative models.
    \item \textit{Human Preference Score v2 (HPS v2)} \citep{wu2023human} is a benchmark designed to evaluate the alignment of T2I models with human preferences. It is built upon Human Preference Dataset v2 (HPD v2), a comprehensive dataset comprising 798k human-annotated pairwise comparisons across nine generative models and real images from the COCO dataset. HPS v2 trains a fine-tuned CLIP-based preference prediction model, demonstrating superior generalization and sensitivity to algorithmic improvements compared to prior metrics such as Inception Score and Fréchet Inception Distance (FID Score).
    \item \textit{InterleavedBench} \cite{liu2024holistic} is a holistic benchmark designed for evaluating the interleaved generation of text and images. It encompasses diverse real-world use cases, including storytelling, marketing, and multimodal script generation, with 815 instances distributed across 10 categories. Unlike prior benchmarks, InterleavedBench emphasizes arbitrary interleaving of text and images within both inputs and outputs. Its evaluation metric, InterleavedEval, powered by GPT-4, assesses five dimensions: text quality, perceptual quality, image coherence, text-image coherence, and helpfulness.
    \item \textit{Video-ChatGPT} \citep{Maaz2023VideoChatGPT} contains a sophisticated framework for evaluating the text generation capabilities of video-based conversational models, leveraging the ActivityNet-200 \citep{caba2015activitynet} dataset's rich collection of videos with dense descriptive captions. This comprehensive evaluation system employs GPT-3.5 to assess model outputs on a 1-5 scale across five fundamental dimensions: Correctness of Information (verifying factual accuracy and alignment with video content), Detail Orientation (examining response completeness and specificity), Contextual Understanding (assessing comprehension of overall video context), Temporal Understanding (evaluating grasp of event sequences), and Consistency (measuring reliability across similar queries). 
    \item \textit{AIR-Bench} \citep{yang2024air} is designed to assess the audio-centric interaction capabilities of Large Audio-Language Models (LALMs). Unlike previous benchmarks that primarily focus on fundamental tasks like automatic speech recognition, AIR-Bench offers a comprehensive evaluation of LALMs' ability to understand and interact with various audio signals, including human speech, natural sounds, and music, in a textual format. It comprises two main components: the foundation benchmark, which includes 19 tasks with approximately 19,000 single-choice questions to test basic single-task abilities, and the chat benchmark, featuring 2,000 instances of open-ended QA data to evaluate complex audio comprehension and instruction-following capabilities, utilizing advanced language models such as GPT-4 for scoring. Since Qwen2-Audio-Instruct is a chat model, our experiment results come from its open-ended QA part.
\end{itemize}

\subsection{Hyper-Parameters}

The hyper-parameters involved in our experiment are as follows. Their selection was based on community implementations \citep{sun2023aligning, zheng2024llamafactory} and our experimental observations. They may not be optimal, but we have confirmed that they do not affect the correctness of the training. These hyper-parameters vary across different modalities, but we ensure consistency between language feedback and binary feedback scenarios.

\begin{table}[ht]
\renewcommand{\arraystretch}{1.0}
\centering
\footnotesize
\resizebox{\columnwidth}{!}{
\begin{tabular}{@{}lccccc@{}}
\toprule
\textbf{Hyperparameters} & \multicolumn{1}{c}{TI2T} & \multicolumn{1}{c}{T2I} & \multicolumn{1}{c}{TI2TI} & \multicolumn{1}{c}{TV2T} & \multicolumn{1}{c}{TA2T} \\ \midrule
Epochs & 3 & 3 & 3 &3  & 3 \\
Batch Size Per Device & 4 & 4 & 4 & 3  & 4 \\
Learning Rate & 1.e-6 & 1.e-6 & 1.e-6 & 1.e-7  & 1.e-6 \\
Scheduler Type & cosine & cosine & cosine & cosine  & cosine \\
Warmup Ratio & 0.03 & 0.03 & 0.03 & 0.1  & 0.03 \\
Gradient Accumulation & 1 & 2 & 2 &1  & 1 \\
Weight Decay & 0.00 & 0.00 & 0.00 & 0.00  & 0.00 \\
Max Token Length & 4096 & 4096 & 4096 & 4096 & 4096 \\
BFloat16 & True & True & True & True  & True \\ \bottomrule
\end{tabular}
}
\vspace{-0.8em}
\caption{\textbf{Hyper-parameters for feedback modeling.}}
\label{tab:llf_feedback_model_hyper}
\end{table}

\begin{table}[ht]
\renewcommand{\arraystretch}{1.0}
\centering
\footnotesize
\resizebox{\columnwidth}{!}{
\begin{tabular}{@{}lccccc@{}}
\toprule
\textbf{Hyperparameters} & \multicolumn{1}{c}{TI2T} & \multicolumn{1}{c}{T2I} & \multicolumn{1}{c}{TI2TI} & \multicolumn{1}{c}{TV2T} & \multicolumn{1}{c}{TA2T} \\ \midrule
Epochs & 3 & 3 & 3 &3 & 3 \\
Batch Size Per Device & 4 & 4 & 4 & 1  & 4 \\
Learning Rate & 1.e-6 & 5.e-6 & 5.e-7 & 1.e-6  & 1.e-6 \\
Scheduler Type & cosine & cosine & cosine & cosine  & cosine \\
Warmup Ratio & 0.03 & 0.03 & 0.03 & 0.01  & 0.03 \\
Gradient Accumulation & 1 & 2 & 2 & 1  & 1 \\
Weight Decay & 0.00 & 0.00 & 0.00 & 0.00  & 0.00 \\
Max Token Length & 2048 & 4096 & 4096 & 4096  & 2048 \\
BFloat16 & True & True & True & True  & True \\ \bottomrule
\end{tabular}
}
\vspace{-0.8em}
\caption{\textbf{Hyper-parameters for reward modeling.}}
\label{tab:llf_reward_model_hyper}
\end{table}

\begin{table}[ht]
\renewcommand{\arraystretch}{1.0}
\centering
\footnotesize
\resizebox{\columnwidth}{!}{
\begin{tabular}{@{}lccccc@{}}
\toprule
\textbf{Hyperparameters} & \multicolumn{1}{c}{TI2T} & \multicolumn{1}{c}{T2I} & \multicolumn{1}{c}{TI2TI} & \multicolumn{1}{c}{TV2T} & \multicolumn{1}{c}{TA2T} \\ \midrule
Epochs & 2 & 2 & 2 &2  & 2 \\
Batch Size Per Device & 4 & 4 & 2 &3  & 4 \\
Learning Rate & 1.e-6 & 5.e-7 & 5.e-7 &1.e-7  & 1.e-6 \\
Scheduler Type & cosine & cosine & cosine & cosine  & cosine \\
Warmup Ratio & 0.03 & 0.03 & 0.03 & 0.1  & 0.03 \\
Gradient Accumulation & 1 & 2 & 2 &1  & 1 \\
Weight Decay & 0.00 & 0.00 & 0.00 & 0.00  & 0.00 \\
Max Token Length & 2048 & 4096 & 4096 & 4096 & 2048 \\
BFloat16 & True & True & True & True  & True \\
Scale Coefficient & 0.10 & 0.10 & 0.10 & 0.10 & 0.10 \\
\bottomrule
\end{tabular}
}
\vspace{-0.8em}
\caption{\textbf{Hyper-parameters for DPO.}}
\label{tab:llf_dpo_hyper}
\end{table}

\begin{table}[ht]
\renewcommand{\arraystretch}{1.0}
\centering
\resizebox{\columnwidth}{!}{
\begin{tabular}{@{}lccccc@{}}
\toprule
\textbf{Hyperparameters} & \multicolumn{1}{c}{TI2T} & \multicolumn{1}{c}{T2I} & \multicolumn{1}{c}{TI2TI} & \multicolumn{1}{c}{TV2T} & \multicolumn{1}{c}{TA2T} \\ \midrule
Epochs & 3 & 3 & 3 & 3  & 3 \\
Batch Size Per Device & 4 & 4 & 4 & 1  & 4 \\
Actor Learning Rate & 1.e-5 & 1.e-5 & 1.e-5 & 5.e-8  & 1.e-6 \\
Actor Scheduler Type & cosine & cosine & cosine & cosine  & cosine \\
Actor Warmup Ratio & 0.03 & 0.03 & 0.03 & 0.03  & 0.03 \\
Critic Learning Rate & 5.e-6 & 5.e-6 & 5.e-6 & 5.e-8  & 5.e-6 \\
Critic Scheduler Type & constant & constant & constant & constant  & constant \\
Critic Warmup Ratio & 0.03 & 0.03 & 0.03 & 0.03  &  0.03 \\
Gradient Accumulation & 1 & 4 & 4 & 1  & 1 \\
Weight Decay & 0.01 & 0.0 & 0.0 & 0.0  & 0.01 \\
Max Token Length & 2048 & 4096 & 4096 & 2048  & 2048 \\
BFloat16 & True & True & True & True  & True \\ 
Sampling Temperature & 1.0 & 0.2 & 0.2 & 1.0  & 1.0 \\
\bottomrule
\end{tabular}
}
\vspace{-0.8em}
\caption{\textbf{Hyper-parameters for PPO.}}
\label{tab:llf_ppo_hyper}
\end{table}

\section{Dataset Card}

\subsection{Dataset Overview}
As the number of modalities increases, current all-modality models encounter significant challenges in effectively following instructions. These challenges include difficulties in comprehending multimodal instructions and generating outputs that align with the intended directives \citep{yin2023survey, yu2024rlhf, majumder2024tango}. While RLHF has demonstrated its effectiveness in addressing such issues for specific modalities, such as text and images \citep{yu2024rlhf}, its applicability to all-modality scenarios remains uncertain.

\begin{figure*}[t]
    \centering
    \includegraphics[width=\textwidth]{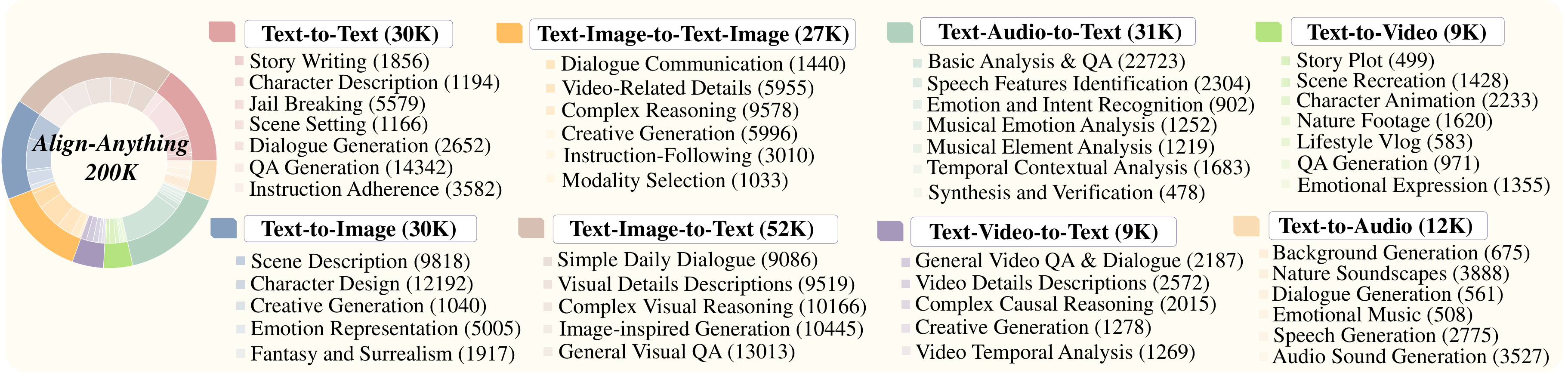}
    \vspace{-1.8em}
    \caption{
    \textbf{Composition of \emph{align-anything-200k} preference dataset.} 
    The composition of enhanced prompts encompasses text, image, video, and audio modalities, aiming to improve the instruction-following capabilities of foundational all-modality models. 
    In addition, We visualize the overall text prompt distribution across all tasks, demonstrating that the data covers different semantic embedding spaces. 
    }
    \label{fig:pie-chart}
\end{figure*}

Furthermore, existing preference datasets predominantly focus on single-modal tasks, lacking the comprehensive information required to capture the intricacies of all-modality features. In response, we introduce \textit{align-anything-200k}, the first all-modality human preference dataset designed to enhance the instruction-following capabilities of all-modality models. This dataset encompasses eight subtasks across text, image, audio, and video modalities (see \cref{fig:pie-chart}). For the examples please refer to the \texttt{dataset/dataset\_examples} included in the supplementary materials.

To ensure consistent modality preference modeling, we categorize targets into \textit{modality-agnostic} and \textit{modality-specific} types, which act as evaluation metrics for instruction-following (see \cref{app:instruction-following-dimensions}). A comprehensive dataset collection and annotation pipeline are outlined in \cref{app:data-annotation-collection}, while \cref{app:statistical-properties-dataset} provides a detailed analysis of dataset features, including prompt semantics, preference features \textit{etc.}. Additionally, examples from the \textit{align-anything-200k} dataset are presented in \cref{app:align-anything-examples}.

\begin{table}[ht]
\renewcommand{\arraystretch}{1.2}
\centering
\begin{threeparttable}
\resizebox{\columnwidth}{!}{
\begin{tabular}{@{}lcc}
\toprule
\textbf{Tasks} & \textbf{Datasets} & \textbf{Models} \\ \midrule
\stackon{}{\stackon{}{T2T}} & \stackon{Evol-Instruct \citep{xu2024wizardlm}}{\stackon{PKU-SafeRLHF \citep{ji2024pku},}{UltraFeedback \citep{cui2023ultrafeedback},}} & \stackon{LLaMA2-7B/13B/70B-Chat \citep{touvron2023llama}}{\stackon{Alpaca-7B, Alpaca2-7B, Alpaca3-8B}{UltraLM \citep{ding2023enhancing}, WizardLM \citep{xu2023wizardlm},}} \\ \midrule
\stackon{}{\stackon{}{TI2T}} & \stackon{LLaVA-Instruct-150K \citep{liu2024improved}}{\stackon{ART500K \citep{mao2017deepart}, MovieNet \citep{huang2020movienet},}{RLHF-V \citep{yu2024rlhf}, ShareGPT4V \citep{chen2023sharegpt4v},}} & \stackon{LLaVA-v1.6-Vicuna-13B \citep{liu2024llava}}{\stackon{LLaVA-v1.5-7B \citep{liu2024improved}, GPT-4o \citep{openai2024gpt4o},}{Llama 3.2-Vision-11B \citep{dubey2024llama},}} \\ \midrule 
\stackon{}{\stackon{}{T2I}} & \stackon{DiffusionDB \citep{wang2022diffusiondb}}{\stackon{MS COCO\citep{chen2015microsoft},}{HPDv2 \citep{wu2023human}, Pick-a-Pic-v2 \citep{kirstain2023pick},}} & \stackon{FLUX.1-schnell, Chameleon-7B \citep{team2024chameleon}}{\stackon{Stable Diffusion v2-1 \citep{Rombach_2022_CVPR},}{Stable Diffusion XL \citep{podellsdxl},}} \\ \midrule 
\stackon{}{TI2TI} & \stackon{LLaVA-Instruct-150K \citep{liu2024improved}}{RLHF-V \citep{yu2024rlhf},} & \stackon{}{Chameleon-7B \citep{team2024chameleon}} \\ \midrule
\stackon{}{\stackon{}{TV2T}} & \stackon{ShareGPT4Video \citep{chen2024sharegpt4video}}{\stackon{NExTQA \citep{xiao2021next}, Panda-70M \citep{chen2024panda}}{BDD100K \citep{yu2020bdd100k}, Ego4D \citep{grauman2022ego4d},}} & \stackon{}{\stackon{Qwen2-VL-7B \citep{Qwen2VL}}{Gemini 1.5 Pro \cite{reid2024gemini},}} \\ \midrule 
\stackon{}{\stackon{}{T2V}} & \stackon{ShareGPT4Video \citep{chen2024sharegpt4video}}{\stackon{mixkit \citep{mixkit},}{VidProm \citep{wang2024vidprom},}} & \stackon{Open-Sora \citep{opensora}, Pika \citep{pika}}{\stackon{CogVideo \citep{hongcogvideo},}{CogVideoX-5B \citep{yang2024cogvideox},}} \\ \midrule
\stackon{}{TA2T} & \stackon{OpenAQA \citep{gong2023listen}, MusicCaps \citep{agostinelli2023musiclm},}{Wavcaps \citep{mei2024wavcaps} GigaSpeech \citep{chen2021gigaspeech}} & \stackon{Qwen2-Audio-7B \citep{Qwen2-Audio}}{Gemini 1.5 Pro \cite{reid2024gemini},} \\ \midrule 
\stackon{}{\stackon{}{T2A}} & \stackon{AudioCaps \citep{kim2019audiocaps}}{\stackon{Audio-Alpaca \citep{majumder2024tango},}{Wavcaps \citep{mei2024wavcaps},}} & \stackon{Stable Audio Open 1.0 \citep{evans2024stable}}{\stackon{Tango 2 \citep{majumder2024tango}, }{AudioLDM 2-large \citep{liu2024audioldm},}} \\ \bottomrule
\end{tabular}
}
\end{threeparttable}
\vspace{-0.8em}
\caption{\textbf{Related datasets and models for each subtask.} We extensively collect datasets of various modalities and enhance the prompt sources for these datasets, resulting in the final prompt sources of \emph{align-anything-200k}. We also widely gather the generation results from both open-source and closed-source models. Llama2-7B and Llama3-8B is fine-tuned with Alpaca-52K\citep{taori2023stanford}, resulting in Alpaca2-7B and Alpaca3-8B.}

\label{tab:task_datasets_models}
\end{table}

\subsection{Instruction-Following Dimensions}
\label{app:instruction-following-dimensions}

In the context of text modality, instruction-following refers to the ability of LLMs to effectively execute human-provided instructions, such as answering questions or summarizing text. This capability enables them to function as helpful, harmless, and honest assistants \citep{ouyang2022training,touvron2023llama}. However, as the number of modalities increases, establishing a unified metric for instruction-following across all modalities becomes increasingly challenging. While the goal is for an all-modality model to effectively perform instructions across various modalities, each specific modality (\textit{e.g.}, video) may have unique requirements, such as ensuring temporal consistency in video outputs.

To address the diversity inherent in the concept of instruction-following within an all-modality alignment setting, we decompose instruction-following into \textit{modality-agnostic} and \textit{modality-specific} dimensions. The modality-agnostic dimensions are universally applicable across all modalities, while the modality-specific dimensions represent preference criteria tailored to the unique characteristics of each modality. For the annotation prompts of each subtask, please refer to \texttt{dataset/annotation\_prompts} included in the supplementary materials.

\subsubsection{Modality-Agnostic Dimensions}

In this section, we provide a detailed overview of modality-agnostic dimensions for evaluating instruction-following in all-modality scenarios.

\paragraph{Prompt adherence} Prompt adherence refers to the extent to which responses align with the given input prompts, accurately reflecting the specified elements, themes, or instructions. This ensures that the output remains faithful to the user's intent and maintains relevance to the multimodal content of prompts.

\paragraph{Rule conformity} Rule conformity refers to the compliance of responses with logical, physical, biological, or scientific principles that are applicable to the scenario or theme described in the prompt. This metric evaluates the response's consistency with established rules and modality-specific constraints, ensuring realism and plausibility.

\paragraph{Information richness} Information richness evaluates the depth and detail provided in responses, emphasizing the thoroughness and comprehensiveness with which the problem or query is addressed. High-quality responses should offer nuanced insights, detailed explanations, and well-rounded coverage of the topic.

\subsubsection{Modaltiy-Specific Dimensions}
In the following part, we provide a detailed overview of the instruction-following standards designed based on the unique characteristics of different modalities, along with the subtasks to which they apply.

\paragraph{Clarity} Clarity refers to the degree to which responses are clear, comprehensible, and easy to understand. It emphasizes overall coherence, language quality, and grammatical correctness, ensuring that the output effectively communicates its intended meaning without ambiguity. This metric applies to subtasks such as T2T, TI2T, TI2TI, TA2T, and TV2T, where textual response quality plays a critical role in achieving successful multimodal communication.

\paragraph{Aesthetics} Aesthetics refers to the evaluation of the visual and auditory appeal of multimodal outputs, including images, videos, and audio. It encompasses the sensory and emotional impact these outputs deliver to the audience. For image-related subtasks (\textit{e.g.}, T2I, TI2TI), aesthetics involves analyzing aspects such as lighting, color harmony, richness of detail, creativity, and the overall emotional resonance or enjoyment the image evokes. In audio subtasks (\textit{e.g.}, T2A), the focus lies on sound quality, emphasizing clarity, smoothness, and the absence of disruptive elements like noise or distortion. For video subtasks (\textit{e.g.}, T2V), aesthetics includes the evaluation of visual appeal and aesthetic quality, taking into account factors like lighting, color harmony, the richness of detail, and the emotional or immersive experience the video provides.

\paragraph{Cross-modal Consistency} Cross-modal consistency refers to the degree of alignment and harmony between all output modalities. For the TI2TI subtask, this involves ensuring that the text and image are consistent in terms of content, style, and message, so the modalities complement each other and produce a coherent, unified output.

\paragraph{Audio Consistency} Audio consistency evaluates the coherence of an audio output’s acoustic qualities, focusing on smooth transitions along the time axis, natural flow, and the absence of abrupt changes. This metric ensures that the audio maintains a steady tone and rhythm, enhancing the listening experience which is applied in the T2A subtask.

\paragraph{Temporal Consistency} Temporal consistency assesses the smoothness of transitions between video frames, ensuring a natural and fluid progression without sudden jumps or erratic changes in motion. This metric is critical for maintaining the visual flow and is applied in the T2V subtask.

\paragraph{Content Coherence} Content coherence measures the semantic and narrative alignment within a video, ensuring that all elements logically work together to deliver a clear and cohesive message. This metric prevents disjointed or illogical segments and is used in the T2V subtask.

\paragraph{Motion Naturalness} Motion naturalness evaluates the realism and fluidity of object and character movements in a video. It ensures that movements adhere to realistic physical laws, resulting in actions that appear natural, smooth, and believable. This metric is applied in the T2V subtask.

\subsection{Details of Dataset Construction}
\label{app:data-annotation-collection}

\subsubsection{Annotation Pipeline Overview}

In this subsection, we detail the annotation pipeline of the dataset. For text and image modalities, we rely on GPT-4o \citep{openai2024gpt4o} to support the human-AI joint annotation process, while for video and audio modalities, we utilize Gemini-1.5-Pro \citep{reid2024gemini}, as these models currently provide the highest-quality multimodal annotations

\paragraph{Data Pairs Collection} To comprehensively cover all-modality tasks and diverse prompt distributions, in constructing \textit{align-anything-200k}, we first collect open-source multimodal datasets, followed by a refined secondary screening of the corresponding text prompts, videos, and audio. Based on this carefully curated multimodal content, we then utilize advanced multimodal models \citep{openai2024gpt4o, reid2024gemini, Qwen2VL, Qwen2-Audio} to enhance prompts to fit with all-modality alignment requirements, ensuring a strong focus on instruction-following capabilities across modalities. We then gather the responses from open-source and API-based models. The related datasets and models are listed in \cref{tab:task_datasets_models}.

\paragraph{Fine-grained Preference Annotation} We apply a detailed preference annotation process to each question-answer pair, combining insights from both GPT-4o and human annotators. The approach ensures a thorough assessment across \textit{modality-agnostic} and \textit{modality-specific} dimensions, with each dimension scored from 0 to 3 based on strict criteria. Annotators also provide justifications and rationales after providing scores, significantly enhancing the consistency and reliability of the annotation process.

\paragraph{Language Feedback Annotation} To refine response quality further, we implement a comprehensive language feedback annotation process. The process involves defining the critique scope for each response, conducting a structured critique, and offering targeted refinement suggestions. Both human annotators and AI models contribute to this feedback, which is organized into cohesive guidance that addresses specific improvement areas considering different modalities. This feedback process captures fine-grained, modality-related preferences and effectively serves as a rich source of natural human preference across all modalities.

\begin{figure}[ht]
    \centering
    \includegraphics[width=\columnwidth]{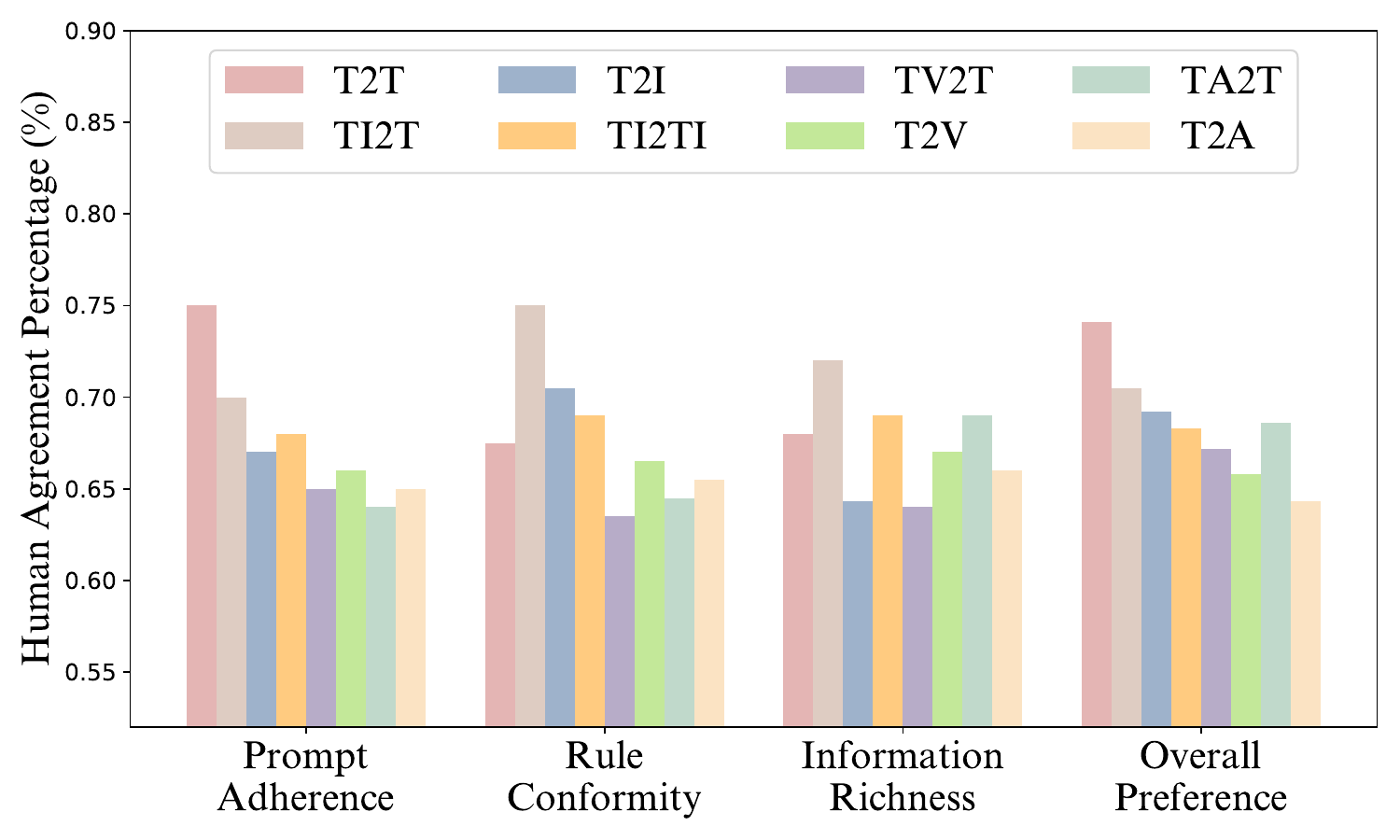}
    \vspace{-2.3em}
    \caption{\textbf{Human-AI preference consistency analysis of \textit{modality-agnostic} dimensions and overall preference.} 
    }
    \label{fig:agreement_appendix_agnostic}
\end{figure}

\subsubsection{Annotation Agreement Analysis}
We evaluate the consistency between AI and human preferences across two dimensions: modality-agnostic and modality-specific. For each subtask, 100 data pairs are randomly selected with 10 annotators for pair-wise comparisons. We assess both preference consistency for modality-agnostic preferences and overall preferences. Additionally, we examine the consistency of modality-specific preferences for each subtask, observing that AI can partially replace human judgment and perform effectively in certain modality-specific tasks. However, our findings reveal that both consistency and accuracy between AI and human preferences decline in multimodal scenarios, suggesting that incorporating multimodal information heightens the complexity of preference judgment. In \cref{fig:agreement_appendix_agnostic}, we illustrate the \textit{modality-agonistic} dimensions consistency across 8 subtasks and overall consistency considering \textit{modality-anostic} and \textit{modality-specific} dimensions.

\subsection{Statistical Properties of Preference Dataset}
\label{app:statistical-properties-dataset}

We analyze the composition of the dataset, including the tasks associated with each modality and the categories of prompts. Based on the characteristics of each modality, we enhance the prompts along instruction-following dimensions. These enhancements cover a variety of types, such as question-answering, multimodal generation, emotional expression, modality selection, and complex reasoning, leveraging advanced multimodal models \citep{openai2024gpt4o, reid2024gemini, Qwen2VL, Qwen2-Audio}. For more details, please refer to \cref{fig:pie-chart}.

\subsection{Examples of Align-anything-200k}
\label{app:align-anything-examples}
\begin{figure*}[h]
\centering
\includegraphics[width=\textwidth]{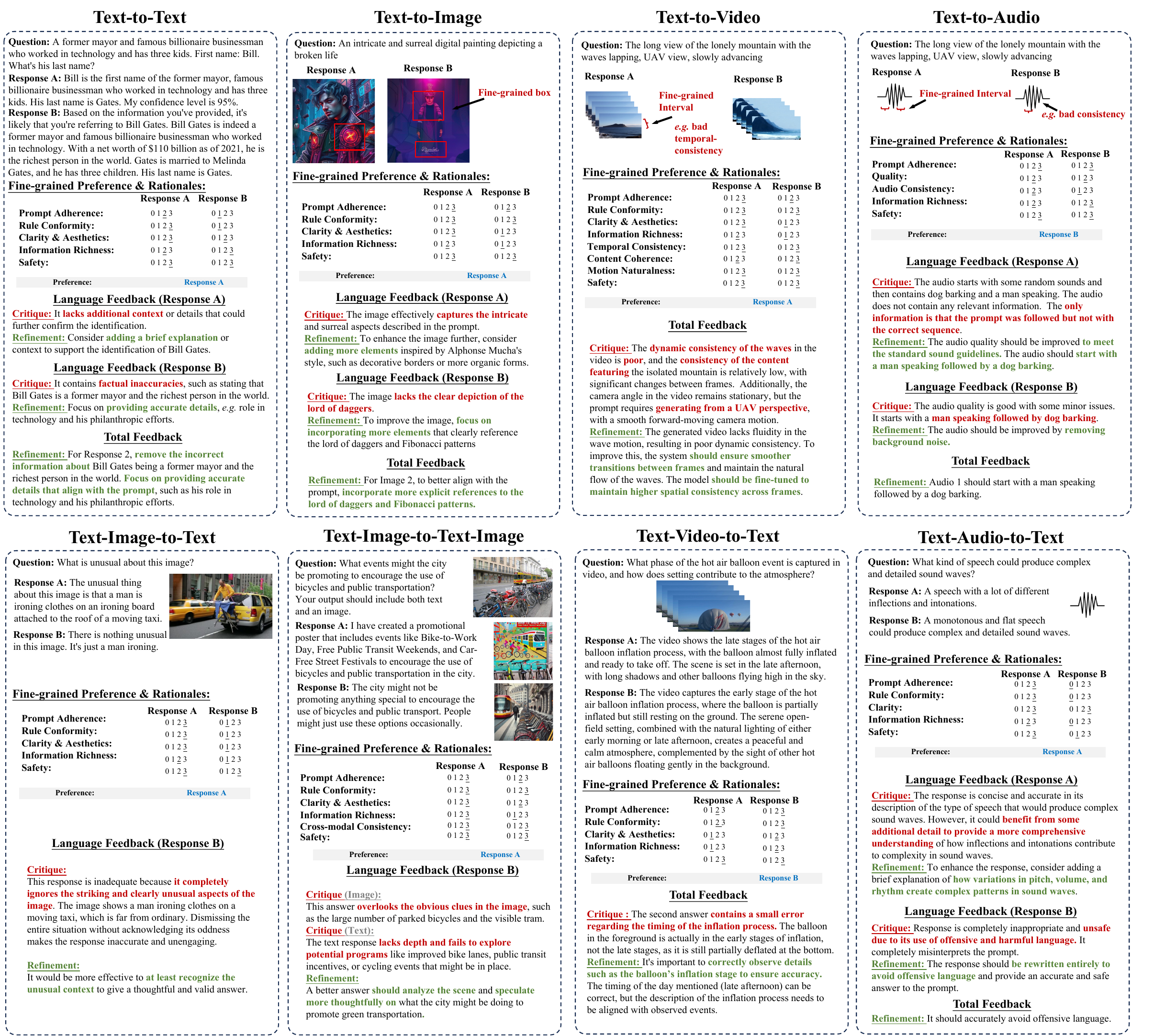}
\caption{\textbf{Examples from the \textit{align-anything-200k} dataset.} We present the training data across 8 subtasks, showcasing fine-grained preferences, evaluation rationales, and detailed language feedback (\textit{critique} and \textit{refinement}) for each response. The feedback highlights modality-specific suggestions and directions for improvement.}
\label{fig:case-demo}
\end{figure*}

In \cref{fig:case-demo}, we showcase the training data across 8 subtasks. For each task, we provide detailed fine-grained preferences and evaluation rationales. Furthermore, we include language feedback annotations for each response. Each response is accompanied by specific critique and refinement feedback, focusing on modality information and directions for improvement. Additionally, the total language feedback further specifies how each response should be revised. For more examples of our dataset, please refer to the \texttt{dataset} folder in the supplementary materials.

\section{Evaluation}

\subsection{All-Modality Understanding}

\subsubsection{Dataset Composition}

The AMU dataset is partially derived from the test set of VGGSound \citep{chen2020vggsound}, supplemented with additional data collected from the internet and generated using generative models. The data is categorized into perception, reasoning, instruction following, and safety, with each category encompassing various types of content. For perception, reasoning, and instruction following, image+audio and video+audio pairs are sourced from VGGSound, while other materials, including manually curated multimedia such as videos, audio, and images, are gathered from the internet. For the safety category, the dataset integrates both internet-sourced content and data generated using generative models.

\subsubsection{Annotation Details}
\begin{figure*}[h]
    \centering
    \includegraphics[width=\textwidth]{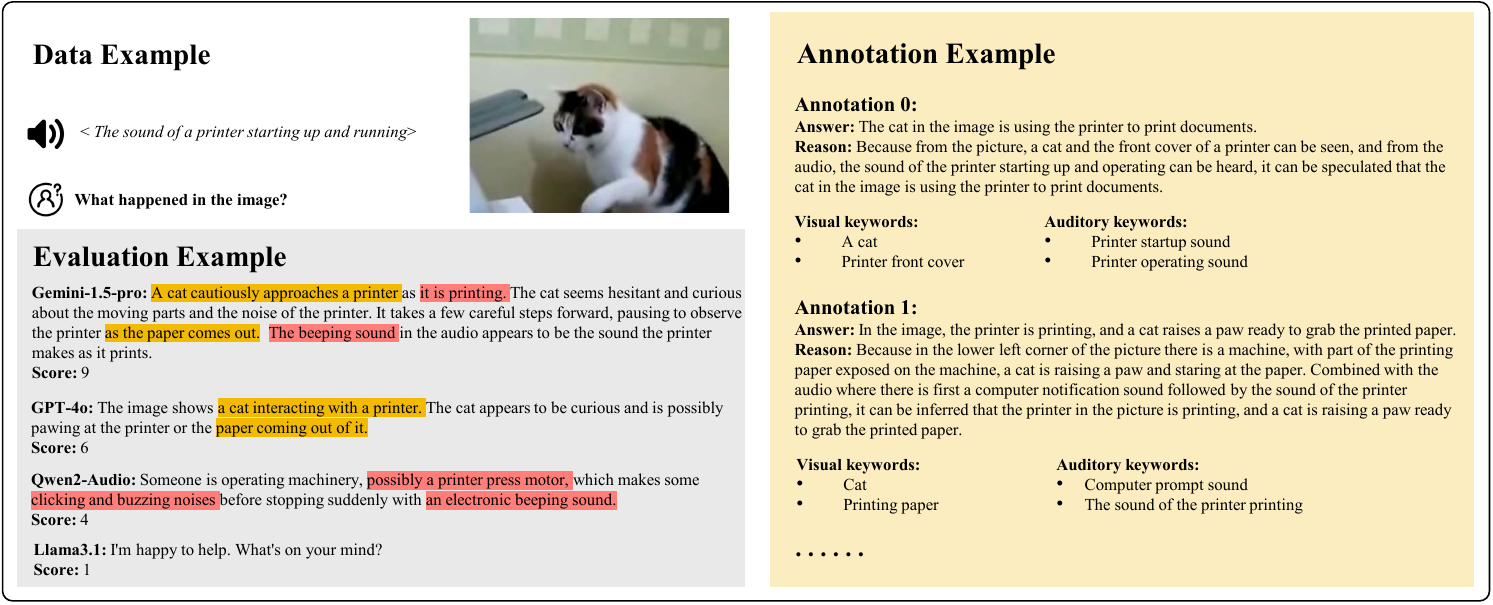}
    \caption{\textbf{Example of AMU task.} The top left corner shows a sample of test data, which includes a piece of visual information, a piece of auditory information, and a question. The right side presents an example of annotations provided by annotators. For an open-ended question, each annotator's response may differ. However, as long as the answer correctly addresses the multimodal information, it will be accepted as a reference answer. The bottom left corner displays the responses of various models to the test data. The more modalities a model can recognize, the more details its reply can incorporate, resulting in higher quality and, consequently, a higher score.}
    \label{fig:example_reasoning}
\end{figure*}

To collect human responses for evaluating the AMU task, we instruct crowdworkers to provide detailed answers and annotations for each test sample.
As shown in \cref{fig:example_reasoning}, each test instance in our dataset consists of one visual data (image or video), one auditory data (audio or speech), and one text question. For video, audio and speech content, we maintain a consistent duration of approximately 10 seconds.
Each test instance is annotated by 10 human annotators who are instructed to provide detailed responses according to the following requirements:

\begin{enumerate}
    \item A direct answer to the text question.
    \item A comprehensive explanation of their reasoning process (minimum 30 characters), detailing how they arrive at their answer.
    \item Keywords extracted from visual and auditory modalities (2 keywords for each modality).
\end{enumerate}

For data in the instruction following category, annotators are required to follow the instructions strictly when providing responses.
In contrast, for data in the safety category, annotators are encouraged to make their own judgments based on their personal values regarding whether the content in the given scenario is appropriate to respond to. If the content is deemed sensitive and unsuitable for a response, annotators should politely decline the request to answer.

\subsubsection{Evaluation Details}
\label{app:amu_eval_prompt}
The default evaluation system prompt consists of two sections. The first section contains instructions for GPT-4, outlining the content it should consider during evaluation and specifying the expected output format. The last section includes reference answers and keywords for each question.

To assess response scores, we guide GPT-4 to evaluate from multiple perspectives, using keywords to determine if it correctly interprets multimodal information. Additionally, multiple manually annotated reference answers are provided to evaluate the alignment between the model's responses and human response distributions.

Due to the unique nature of the safety and instruction following components, we design additional system prompts to evaluate response safety and instruction following. For safety evaluation, GPT-4 should check whether it needs to refuse to answer based on the distribution of human references. For instruction following, GPT-4 should examine whether the response follows the special instructions in the test cases by checking if the reference answers have a special output format. All instruction-following related guidelines and evaluation methods are referenced from IFEval \citep{zhou2023instruction} and FollowBench \citep{jiang-etal-2024-followbench} respectively. The prompts for AMU task are shown in the \texttt{evaluation/prompt} directory included in the supplementary materials.

\subsection{All-Modality Generation}

Due to the limitations of current all-modal generation models, we employs LLMs as agents to perform multimodal tool calls during experiments, allowing models to generate outputs in non-textual modalities. The prompt for LLM agents is shown in the \texttt{evaluation/prompt} directory included in the supplementary materials.

\subsubsection{Instruction Following}
To evaluate the instruction-following capability of multimodal models in different modality generation tasks, we design a pipeline for evaluating instruction-following across various modalities. For the four modalities—text, image, video, and audio—we consider various tasks and dimensions of instruction-following and meticulously design 100 instruction-following prompts for each modality.
During the evaluation phase of the generated content, since different modalities have different evaluation models, we use GPT-4 \cite{achiam2023gpt} and GPT-4o \cite{openai2024gpt4o} for text and image modalities to directly assess the instruction-following level of the generated content using a scalar score from 0 to 10. For audio and video modalities, inspired by the TIFA \cite{hu2023tifa} pipeline, we design multiple-choice questions closely related to the instruction to capture the information in the prompts in more detail. We use Qwen2-Audio \cite{Qwen2-Audio} and Qwen2-VL \cite{Qwen2VL} as evaluation models to assess the generated audio and video, scoring based on the accuracy of the multiple-choice answers, with scores ranging from 0 to 10. 
The system prompts involved in the entire scoring process can be found in the \texttt{evaluation/prompt} directory included in the supplementary materials

\subsubsection{Modality Selection}
To assess the model’s flexibility in choosing different modality combinations based on textual instructions, we engage human crowdworkers to annotate the expected output modality for each of the 100 prompts. Specifically, each instruction is annotated through multiple-choice selections by 25 human annotators across options like text, image, audio, and text, video, and audio. This process produces a distribution of votes for each instruction, reflecting preferences for the expected output modalities. Modality combinations that closely align with human preferences earn additional points in the modality selection metric. 
The human annotation criteria are as follows:

\begin{itemize}[left=0.2cm]
    \item \textbf{Information Richness:} Annotators should specify the desired modality of the model output for a given textual instruction to maximize information gain in the response.

    \item \textbf{Necessity \& Conciseness:} When choosing modality combinations, necessity and conciseness should guide the selection. If a modality adds no significant information, it should be excluded based on the principle of conciseness.
\end{itemize}

\label{app:judge_model}
Modality synergy is a complex attribute, and currently, no simple or objective method exists to uniformly evaluate information across different modalities. While human evaluation or the use of AI under strictly defined annotation guidelines are feasible solutions, they introduce additional costs and complexities. Therefore, we meticulously construct training data for the Modality Synergy task and train a judge model. The judge model can evaluate the synergy metrics of multi-modal information generated according to the task instructions.
\begin{itemize}[left=0.2cm]
\item \textbf{Data Collection}
    Due to the current immaturity of end-to-end multimodal generation models, generating high-quality training data remains a challenge. To address this, we develop a data generation system centered on an LLM Agent. Upon receiving user instructions, the agent calls image, audio, or video generation models through tool usage, producing the corresponding modality information and responding to the user’s instructions in conjunction with text.
\item \textbf{Data Annotation}
    As there is currently no fully automated annotation tool for all modalities information, we develop a comprehensive human annotation pipeline specifically for the modality synergy task. Human annotators receive data pairs consisting of one instruction and outputs from two models, where the model outputs involve any two modalities in text, image, video, audio. Annotators evaluate the two responses based on relevance and consistency, assigning preference annotations in line with the annotation guidelines. Detailed annotation guidelines. please refer to \cref{sec:annotate_modality_synergy}.
    
\item \textbf{RM Training} 
    Referring to reward model training in the text-only modality, the base model of our judge model is Llama3-8B, equipped with a score head.  ImageBind \citep{girdhar2023imagebind} and independent projectors for each modality enable the model to achieve multimodal understanding. The training process has two stages: first, the parameters of the LLM and ImageBind are frozen, and only the projector is trained; second, both the LLM and projector are updated simultaneously.
\end{itemize}

\subsubsection{More Experiment Results}

To evaluate agent performance across various multimodal generative models, we conduct supplementary experiments. Specifically, the LLM-based agent invoke Stable-Diffusion-v1-5 \citep{Rombach_2022_CVPR}, AudioLDM-Small-full-v2 \citep{liu2023audioldm}, and CogVideoX-2B \citep{yang2024cogvideox} for image, audio, and video generation, respectively.  The results of these experiments are presented in \cref{tab:more_eval_results_video} and \cref{tab:more_eval_results_image}.

The two tables align with the conclusions in Section 4.2, emphasizing that current models have limitations, and no single model exhibits a clear advantage in the AMG task. As for the instruction-following metric, the performance of close-sourced llm-based agent declines significantly when the capability of tool models decrease. Analysis of specific test cases shows that close-sourced models often generate more fine-grained instructions for identical test prompts compared to the open-sourced model. Stronger tool models effectively use these fine-grained instructions to produce outputs that meet the test requirements. In contrast, weaker tool models struggle with overly complex and detailed instructions, leading to reduced output quality and lower instruction-following scores. This issue is especially evident in tasks such as video and audio generation, which rely heavily on temporal sequencing. The two closed-source models often offer instructions refined to the temporal dimension. However, the corresponding video and audio generation models are unable to fully adhere to the required temporal sequences, leading to the loss of critical information. Conversely, open-sourced models, through more generalized instructions, better guide tool models to produce outputs closely follow with the test instructions.

\begin{table*}[ht]
\centering
\renewcommand{\arraystretch}{1.2}
\resizebox{1.0\textwidth}{!}{
\begin{tabular}{lcccccccccccccc}
    \toprule
     \multirow{3}{*}{\textbf{Initial Models}} & \multicolumn{5}{c}{\textbf{All-Modality Understanding}} & \multicolumn{8}{c}{\textbf{All-Modality Generation}}& \multirow{3}{*}{\textbf{Overall}}\\
     \cline{2-6}
     \cline{7-14}
     ~ & \multicolumn{4}{c}{\textbf{Category}} & \multirow{2}{*}{\parbox{0.8cm}{\centering \textbf{AMU} \textbf{Score}}} & \multirow{2}{*}{\parbox{1.2cm}{\centering \textbf{Modality} \textbf{Select}}} & \multicolumn{3}{c}{\textbf{Instruction Following}}& \multicolumn{3}{c}{\textbf{Modality Synergy}} & \multirow{2}{*}{\parbox{0.8cm}{\centering \textbf{AMG} \textbf{Score}}} & \\
     \cline{2-5}
     \cline{8-10}
     \cline{11-13}
     ~ & Perception & Reasoning & IF & Safety &  &  & T & V & A & T-V & T-A & V-A &  & \\
     \midrule
     LLaVA-v1.5-7B$^\dag$ & 2.66 &2.67 & 2.50 &2.90 &2.68 & 0.153 & 5.62 & \textbf{8.00} & 4.73 & 0.33 & 0.37 & 0.67 & 1.29 & 1.99 \\
     Qwen2-VL$^\dag$ & 2.76 & 3.07 &2.40 &4.05 &3.07 & 0.143 & 6.70 & 7.97 & 4.83 & 0.55 & 0.59 & 0.68 & 1.69 & 2.38 \\
     Qwen2-Audio$^\dag$ & 3.58 &4.53 &3.40 &2.65 &3.54 & 0.135 & 5.93 & 7.95 & 4.85 & 0.35 & 0.38 & 0.67 & 1.18 & 2.36 \\
     Chameleon-7B$^\dag$ & 1.44 &2.97 &2.80 &2.45 &2.41 & 0.153 & 4.21 & 7.71 & 4.66 & 0.60 & 0.62 & \textbf{0.69} & 1.62 & 2.02 \\
     Llama3.1-8B-Instruct$^\dag$ & 1.05 &1.20 &1.20 &1.35 &1.20 & \textbf{0.212} & 7.69 & 7.87 & 5.15 & \textbf{0.72} & \textbf{0.75} & 0.67 & 3.13 & 2.17 \\
     Gemini-1.5-Pro$^\dag$ & \textbf{5.36} & \textbf{5.67} &\textbf{6.70} & \textbf{6.70} & \textbf{6.11} & 0.193 & 9.45 & 7.95 & \textbf{6.76} & 0.70 & 0.72 & 0.68 & \textbf{3.27} & \textbf{4.69}\\
     GPT-4o$^\dag$ & 2.66 &3.48 &4.20 &5.15 &3.87 & 0.163 & \textbf{9.51} & 7.91 & 6.64 & 0.55 & 0.57 & 0.67 & 2.33 & 3.10 \\
    \bottomrule
\end{tabular}
}
\vspace{-0.5em}
\caption{\textbf{The performance of models in the eval-anything benchmark.} The models performance in AMU task is the same as \cref{tab:eval_main_results}. (\textbf{$\dag$}) indicates that models are used as agents to invoke \textbf{AudioLDM2-Large} \cite{liu2024audioldm} and \textbf{CogVideoX-2B} \cite{yang2024cogvideox} for audio and video generation.
}
\label{tab:more_eval_results_video}
\vspace{-1.0em}
\end{table*}

\begin{table*}[ht]
\centering
\renewcommand{\arraystretch}{1.2}
\resizebox{1.0\textwidth}{!}{
\begin{tabular}{lcccccccccccccc}
    \toprule
     \multirow{3}{*}{\textbf{Initial Models}} & \multicolumn{5}{c}{\textbf{All-Modality Understanding}} & \multicolumn{8}{c}{\textbf{All-Modality Generation}}& \multirow{3}{*}{\textbf{Overall}}\\
     \cline{2-6}
     \cline{7-14}
     ~ & \multicolumn{4}{c}{\textbf{Category}} & \multirow{2}{*}{\parbox{0.8cm}{\centering \textbf{AMU} \textbf{Score}}} & \multirow{2}{*}{\parbox{1.2cm}{\centering \textbf{Modality} \textbf{Select}}} & \multicolumn{3}{c}{\textbf{Instruction Following}}& \multicolumn{3}{c}{\textbf{Modality Synergy}} & \multirow{2}{*}{\parbox{0.8cm}{\centering \textbf{AMG} \textbf{Score}}} & \\
     \cline{2-5}
     \cline{8-10}
     \cline{11-13}
     ~ & Perception & Reasoning & IF & Safety &  &  & T & V & A & T-V & T-A & V-A &  & \\
     \midrule
     LLaVA-v1.5-7B$^\dag$ & 2.66 &2.67 & 2.50 &2.90 &2.68 & 0.182 & 5.62 & 6.78 & \textbf{5.31} & 0.32 & 0.38 & \textbf{0.68} & 1.49 & 2.08 \\
     Qwen2-VL$^\dag$ & 2.76 & 3.07 &2.40 &4.05 &3.07 & 0.177 & 6.70 & 7.06 & 5.14 & 0.60 & 0.62 & 0.66 & 2.09 & 2.58 \\
     Qwen2-Audio$^\dag$ & 3.58 &4.53 &3.40 &2.65 &3.54 & 0.190 & 5.93 & 6.78 & 5.02 & 0.53 & 0.62 & 0.67 & 2.04 & 2.79 \\
     Chameleon-7B$^\dag$ & 1.44 &2.97 &2.80 &2.45 &2.41 & 0.156 & 4.21 & 6.85 & 4.81 & 0.55 & 0.60 & 0.65 & 1.49 & 1.95 \\
     Llama3.1-8B-Instruct$^\dag$ & 1.05 &1.20 &1.20 &1.35 &1.20 & 0.231 & 7.69 & 6.76 & 5.21 & \textbf{0.76} & \textbf{0.79} & 0.67 & 3.37 & 2.29 \\
     Gemini-1.5-Pro$^\dag$ & \textbf{5.36} & \textbf{5.67} &\textbf{6.70} & \textbf{6.70} & \textbf{6.11} & 0.227 & 9.45 & \textbf{7.52} & 4.93 & 0.48 & 0.52 & 0.67 & 2.70 & \textbf{4.41} \\
     GPT-4o$^\dag$ & 2.66 &3.48 &4.20 &5.15 &3.87 & \textbf{0.266} & \textbf{9.51} & 6.77 & 5.02 & 0.70 & 0.73 & 0.67 & \textbf{3.98} & 3.92 \\
\bottomrule
\end{tabular}
}
\vspace{-0.5em}
\caption{\textbf{The performance of models in the eval-anything benchmark.} The models performance in AMU task is the same as \cref{tab:eval_main_results}. (\textbf{$\dag$}) indicates that models are used as agents to invoke \textbf{AudioLDM-Small-full-v2} \cite{liu2023audioldm} and \textbf{Stable-Diffusion-v1-5} \cite{Rombach_2022_CVPR} for audio and image generation.
}
\label{tab:more_eval_results_image}
\vspace{-1.0em}
\end{table*}

\begin{figure*}[h]
    \centering
    \includegraphics[width=\textwidth]{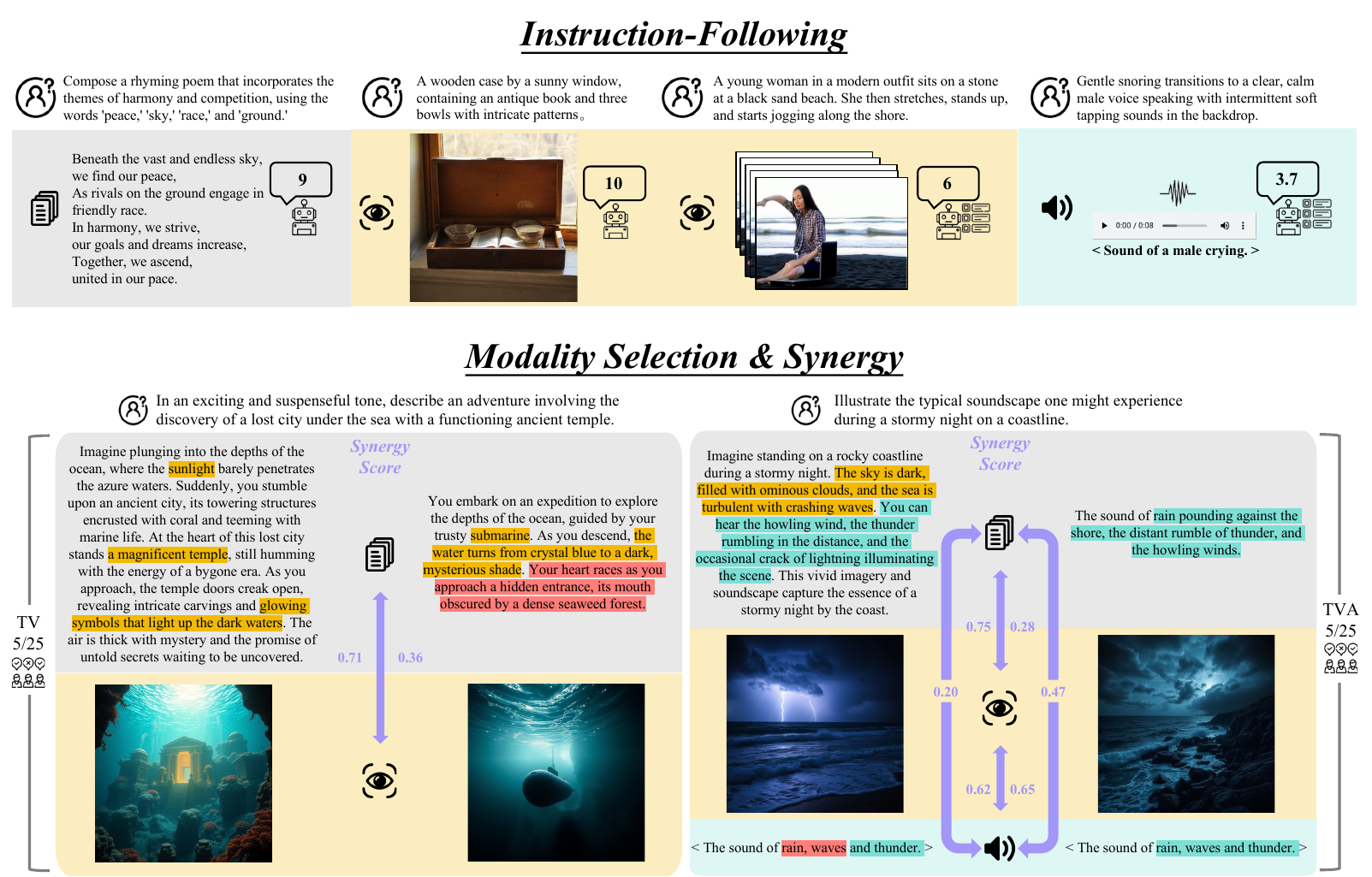}
    \caption{\textbf{Examples of AMG task.} We create two distinct instruction test sets for evaluating the three metrics of the AMG task, dividing the instruction-following section into four modality-based subsets. The rationale is that in multimodal scenarios, evaluating instruction-following requires specific instructions to assess the model’s ability to generate content that aligns with the instructions. Conversely, for modality selection and modality synergy tasks, the instructions should enable open-ended responses to assess the model’s autonomous use of modalities. For detailed example analysis, please refer to \cref{sec:amg_example}}
    \label{fig:amg_example}
\end{figure*}

\subsubsection{Human Annotation for Modality Synergy Task}
\label{sec:annotate_modality_synergy}
This guide outlines the process for evaluating multimodal outputs generated by two different models. Your task is to assess the correlation and consistency between the outputs of \texttt{\{modality\_1\}} and \texttt{\{modality\_2\}} for each model and assign a score to reflect how effectively these outputs complement one another. When scoring, it is important to distinguish between the two models’ outputs, avoiding identical scores for both models unless absolutely necessary.
\paragraph{Evaluation Criteria}
\begin{itemize}[left=0.2cm]
    \item \textbf{Correlation} Evaluate whether the outputs from different modalities are clearly related to the same topic or object.
    \item \textbf{Consistency} Ensure that the information presented across modalities is coherent and does not contradict itself, maintaining uniformity in description or explanation.
\end{itemize}
\paragraph{Scoring Guidelines}
\begin{itemize}[left=0.2cm]
    \item \textbf{Score 5:} The outputs reveal clear contradictions or conflicts between modalities. Either the modalities focus on entirely different topics or objects, or they describe the same topic in an incompatible manner. Such inconsistencies undermine the coherence and trustworthiness of the message, creating confusion.
    
    \item \textbf{Score 4:} The outputs relate to the same topic or object but address independent aspects or dimensions. Although there is no direct contradiction, the modalities fail to complement or overlap meaningfully, resulting in a weak connection.

    \item \textbf{Score 3:} The outputs display basic correlation and consistency. Both modalities describe the same topic or object but do not offer additional insights or value when combined. The information is redundant, and no significant gain is observed from their interaction.

    \item \textbf{Score 2:} The outputs demonstrate strong correlation and consistency, addressing the same topic or object. While minor variations in presentation may exist, they do not conflict but instead provide additional perspectives that moderately enhance understanding. The interaction between modalities is effective, though not as impactful as in the 5-point case.

    \item \textbf{Score 1:} The outputs exhibit an exceptional level of correlation and consistency. Each modality significantly enhances the other, providing unique yet complementary information. Together, they form a cohesive and synergistic combination, offering insights beyond the sum of their individual contributions.
\end{itemize}
\paragraph{Important Notes}
\begin{itemize}
    \item Focus your evaluation on the primary content of each modality, disregarding minor imperfections such as small image distortions, slight spelling errors, or minor audio glitches.
    \item If the outputs contain factual inaccuracies or irrelevant information, score them based on the correlation between the modalities, but cap the maximum score at 4 points (since no meaningful information gain is possible).
    \item The prompt provides contextual background information and should not influence your assessment.
    \item Your evaluation should be limited to the interplay between \texttt{\{modality\_1\}} and \texttt{\{modality\_2\}}, excluding considerations of other modalities.
\end{itemize}

\subsubsection{Evaluation Examples}
\label{sec:amg_example}
Examples of modality synergy evaluation in AMG task refer to \cref{fig:amg_example}.

In the \textit{Instruction-Following} module, we evaluate the model's ability to follow instructions across text, visual (images and videos), and audio modalities. For the text and image modalities, the generated content aligns well with the requirements in the instructions, resulting in high scores in the evaluation. However, the generated content fails to match the instructions for the video and audio modalities. For example, the video does not depict the action of a woman standing up, and the audio does not reflect the sound transition process. As a result, the model can only correctly complete part of the multiple-choice questions related to the original instructions, leading to lower scores.

In the left part of the \textit{Modality Selection \& Synergy} module, the text on the left describes elements such as \textit{sunlight}, \textit{a magnificent temple}, and \textit{glowing symbols}, all of which are effectively mirrored in the corresponding image. In contrast, the image on the right fails to depict key features like \textit{the hidden entrance and seaweed forest} mentioned in the text. This highlights the superior modality synergy of the left response.

In the right part of \textit{Modality Selection \& Synergy} module, the synergy between text and image modalities shows that the left text aligns with its corresponding image, while the right image depicts a similar scene to the left. However, the lack of visual description in the right text creates a significant disparity in their modality synergy score. In terms of text and audio synergy, both responses feature nearly identical audio content. However, the absence of rain and wave sound descriptions in the left text results in a lower score compared to the right. Regarding image and audio synergy, the information conveyed by the images and audio is nearly identical in both responses, resulting in comparable scores.

\end{document}